\documentclass{article}

\PassOptionsToPackage{numbers, compress}{natbib}

\usepackage[preprint]{neurips_2023}

\usepackage[utf8]{inputenc} 
\usepackage[T1]{fontenc}    
\usepackage{hyperref}       
\usepackage{url}            
\usepackage{booktabs}       
\usepackage{amsfonts}       
\usepackage{nicefrac}       
\usepackage{microtype}      
\usepackage{xcolor}         

\newcommand{\bluegain}[1]{\textbf{\frenchblue{(+#1)}}}

\usepackage{graphicx}
\usepackage{booktabs}
\usepackage{amsmath}
\usepackage{amssymb}
\usepackage{amsfonts}
\usepackage{multirow}
\usepackage{verbatim}
\usepackage{caption}
\usepackage{longtable}
\usepackage{supertabular}
\usepackage{float}

\usepackage{enumitem}
\usepackage{tablefootnote}
\usepackage{xcolor}
\usepackage{xspace}
\usepackage{textcomp}
\usepackage{makecell}
\usepackage{multirow}
\usepackage{lscape} 
\usepackage{siunitx}

\usepackage{amsmath}
\usepackage{amsfonts}
\usepackage{amssymb}
\usepackage{wrapfig}
\usepackage{subcaption}
\usepackage{multirow}
 \usepackage{mathtools}

\usepackage{amssymb}%
\usepackage{pifont}%
\usepackage{scrextend}

\usepackage{array}
\usepackage{latexsym}
\usepackage{microtype}
\definecolor{mydarkblue}{rgb}{0,0.08,0.45}
\usepackage{url}            %
\usepackage{nicefrac}       %
\usepackage{changepage}
\usepackage{xargs}          %
\usepackage{wrapfig,lipsum,booktabs}
\usepackage{longtable}
\usepackage{subcaption}
\usepackage{endnotes}

\usepackage{pgfplots}
\usetikzlibrary{pgfplots.groupplots}
\usepackage{tikz}
\usetikzlibrary{patterns}

\usepackage[most]{tcolorbox}

\usepackage{multicol}

\definecolor{battleshipgrey}{rgb}{0.3, 0.3, 0.3}
\definecolor{brilliantrose}{rgb}{1.0, 0.33, 0.64}
\definecolor{americanrose}{rgb}{1.0, 0.01, 0.24}
\definecolor{jweigreen}{rgb}{0,0.45,0.24}
\definecolor{bluegray}{rgb}{0.1, 0.1, 0.4}
\definecolor{ao(english)}{rgb}{0.0, 0.5, 0.0}
\definecolor{blanchedalmond}{rgb}{1.0, 0.92, 0.8}
\definecolor{atomictangerine}{rgb}{1.0, 0.6, 0.4}
\definecolor{chocolate(web)}{rgb}{0.82, 0.41, 0.12}
\definecolor{bananayellow}{rgb}{1.0, 0.88, 0.21}
\definecolor{goldenbrown}{rgb}{0.6, 0.4, 0.08}
\definecolor{aliceblue}{rgb}{0.94, 0.97, 1.0}
\definecolor{beige}{rgb}{0.96, 0.96, 0.86}
\definecolor{babyblue}{rgb}{0.54, 0.81, 0.94}
\definecolor{camel}{rgb}{0.76, 0.6, 0.42}
\definecolor{cinnamon}{rgb}{0.82, 0.41, 0.12}
\definecolor{deepskyblue}{rgb}{0.0, 0.75, 1.0}
\definecolor{frenchblue}{rgb}{0.0, 0.45, 0.73}
\definecolor{classicrose}{rgb}{0.98, 0.8, 0.91}
\definecolor{frenchrose}{rgb}{0.96, 0.29, 0.54}
\definecolor{frenchlilac}{rgb}{0.53, 0.38, 0.56}
\definecolor{frenchbeige}{rgb}{0.65, 0.48, 0.36}

\newcommand{\frenchblue}[1]{{\color{frenchblue}{#1}}}

\usepackage{verbatim}

\usepackage{anyfontsize}

\usepackage{microtype}
\usepackage{graphicx}
\usepackage{booktabs} 
\usepackage{multirow}
\usepackage{amsmath,amssymb}
\usepackage{booktabs}
\usepackage{caption,subcaption}

\usepackage{xcolor}
\definecolor{mygreen}{HTML}{3cb44b}
\definecolor{skyblue}{HTML}{beffff}
\definecolor{lightgreen}{HTML}{90ee90}

\usepackage{tcolorbox}
\usepackage{enumitem}

\newcommand{\palm}[0]{PaLM}

\newcommand{\flanpalm}[0]{\textsc{Flan-PaLM}}

\usepackage{adjustbox}

\newcommand{\RN}[1]{%
	\textup{\lowercase\expandafter{\it \romannumeral#1}}%
}
\usepackage{tabu}

\newcommand{\beq}{\vspace{0mm}\begin{equation}}
\newcommand{\eeq}{\vspace{0mm}\end{equation}}
\newcommand{\beqs}{\vspace{0mm}\begin{eqnarray}}
\newcommand{\eeqs}{\vspace{0mm}\end{eqnarray}}
\newcommand{\barr}{\begin{array}}
\newcommand{\earr}{\end{array}}

\usepackage{color, colortbl}
\definecolor{Gray}{gray}{0.93}

\usepackage{lipsum}

\usepackage{xcolor,amsmath}
\usepackage[linesnumbered,ruled,vlined]{algorithm2e}
\DontPrintSemicolon

\usepackage{color, colortbl}

\definecolor{emerald}{rgb}{0.31, 0.78, 0.37}

\newcommand{\MyColorBox}[2][red]%
{%
    \settowidth{\Width}{#2}%
    \colorbox{#1}%
    {%
        \raisebox{-\DepthReference}%
        {%
                \parbox[b][\HeightReference+\DepthReference][c]{\Width}{\centering#2}%
        }%
    }%
}

\definecolor{codegray}{gray}{0.9}

\SetKwComment{Comment}{\color{green!50!black}\# }{}

\SetKwProg{Function}{def}{:}{}

\SetKwProg{For}{for}{:}{}
\SetKwProg{If}{if}{:}{}

\newcommand{\shortname}{\textsc{Flan-MoE}}
\newcommand{\sshortname}{\textsc{Flan}}

\usepackage{pifont}
\title{Mixture-of-Experts Meets Instruction Tuning:  \\ A Winning Combination for Large Language Models}
\author{
\textbf{\normalsize{Sheng Shen$^{\natural *}$~~~~Le Hou$^{\dagger}$~~~~Yanqi Zhou$^{\dagger}$~~~~Nan Du$^{\dagger}$~~~~Shayne Longpre$^{\top *}$~~~~Jason Wei$^{\dagger}$}}, \\
\\
\textbf{\normalsize{Hyung Won Chung$^{\dagger}$~~~~Barret Zoph$^{\dagger}$~~~~William Fedus$^{\dagger}$~~~~Xinyun Chen$^{\dagger}$~~~~Tu Vu$^{\ddag *}$}}, \\
\\
\textbf{\normalsize{Yuexin Wu$^{\dagger}$~~~~Wuyang Chen$^{\S *}$~~~~Albert Webson$^{\dagger}$~~~~Yunxuan Li$^{\dagger}$~~~~Vincent Zhao$^{\dagger}$~~~~Hongkun Yu$^{\dagger}$}} \\
\\
\textbf{\normalsize{Kurt Keutzer$^{\natural}$~~~~Trevor Darrell$^{\natural}$~~~~Denny Zhou$^{\dagger}$}} \vspace{0.3cm} \\
\\
\normalsize{$^{\dagger}$Google~~~~~~$^{\natural}$University of California, Berkeley~~~~~~$^{\top}$Massachusetts Institute of Technology} \vspace{0.3cm} \\
\normalsize{$^{\ddag}$University of Massachusetts Amherst~~~~~~$^{\S}$The University of Texas at Austin}
}

\begin{document}

\maketitle
\let\thefootnote\relax\footnotetext{\hspace*{-0.5cm}* Work done at Google}

\begin{abstract}
    Sparse Mixture-of-Experts (MoE) is a neural architecture design that can be utilized to add learnable parameters to Large Language Models (LLMs) without increasing inference cost. Instruction tuning is a technique for training LLMs to follow instructions. We advocate combining these two approaches, as we find that MoE models benefit more from instruction tuning than dense models. In particular, we conduct empirical studies across three experimental setups: (i) Direct finetuning on individual downstream tasks devoid of instruction tuning; (ii) Instruction tuning followed by in-context few-shot or zero-shot generalization on downstream tasks; and (iii) Instruction tuning supplemented by further finetuning on individual downstream tasks. In the first scenario, MoE models overall underperform dense models of identical computational capacity. This narrative, however, dramatically changes with the introduction of instruction tuning (second and third scenario), used independently or in conjunction with task-specific finetuning.
    Our most powerful model, \shortname{}$_\textsc{32b}$, surpasses the performance of \flanpalm$_\textsc{62b}$ on four benchmark tasks, while using only a third of the FLOPs. The advancements embodied by \shortname{} inspire a reevaluation of the design principles of large-scale, high-performance language models in the framework of task-agnostic learning.
\end{abstract}
\section{Introduction}
\label{sec:introduction}

The recent years have witnessed remarkable advancements in the field of natural language processing (NLP), driven by the development of increasingly large and sophisticated deep learning models. Among these models, transformer-based language models \cite{transformer} have emerged as the de facto standard for a wide range of NLP tasks, owing to their unparalleled capabilities in capturing complex linguistic patterns and generalizing across diverse contexts. 
One particularly successful paradigm for training such models is instruction-tuning~\cite{t0,flan,flant5,longpre2023flan,bloomz,gptrlhf}, which enhances their performance on specific tasks by adapting their pre-trained representations to follow natural language instructions.

While the benefits of Large Language Models (LLMs) are indisputable, their rapidly growing size and computational requirements pose significant challenges in terms of training efficiency, memory footprint, and deployment costs. Consequently, there is a pressing need for developing scalable techniques that can harness the power of these models without incurring prohibitive computational overheads.

On the other hands, models with sparsely activated Mixture of Experts (MoEs) significantly reduce the computational cost of LLMs. MoE models build upon the observation that language models can be decomposed into smaller, specialized sub-models, or "experts", that focus on distinct aspects of the input data, thereby enabling more efficient computation and resource allocation. However, we show that conventional, task-specific finetuning MoE models lead to suboptimal performance, often even worse than finetuning dense models with the same computational cost. One of the possible reasons is the discrepancy between general pretraining and task-specific finetuning.

In this paper, we illuminate the pivotal role of instruction-tuning within the context of Mixture-of-Experts (MoE) models, specifically in terms of their successful scalability on downstream tasks. We demonstrate this through a two-fold analysis: Firstly, we expand on the known benefits of instruction-tuning for task-specific downstream finetuning~\cite{longpre2023flan}, illustrating its significantly larger impact when applied to MoE models compared to their dense equivalents. Secondly, we emphasize the necessity of an instruction-tuning stage for MoE models~\cite{moe,glam,switchtransformer,gshard} to surpass the performance of dense models on downstream and held-out tasks. 
Our unique amalgamation, \shortname{}, is an instruction-tuned model built on the Flan mixture\cite{flant5}, which successfully harnesses the strengths of both instruction-tuning and the sparse MoE technique. \shortname{} effectively and efficiently scales up language models, without necessitating a rise in computational resources or memory requirements.


We subject our model, \shortname{}, to a battery of tests across an array of tasks encompassing natural language understanding, reasoning, and question answering. Our evaluation framework consists of three distinct setups: (i) Direct finetuning of the model on individual downstream tasks; (ii) Instruction tuning succeeded by in-context, few-shot, or zero-shot generalization on downstream tasks; and (iii) Instruction tuning enhanced with subsequent finetuning on individual downstream tasks. 
The results spotlight \shortname{}'s marked superiority over its dense counterparts in the second and third settings. Notably, these advancements materialize without the need for augmented computational resources or memory requisites. Our top-tier model, in fact, manages to eclipse the performance of a \flanpalm~equivalent, requiring only a third of the computational cost per token on four separate benchmarks.

To summarize, our contributions are as follows:
\begin{itemize}
    \item We establish the critical role of instruction-tuning in the efficacy of MoE models:
    \begin{itemize}
        \item We demonstrate that in the absence of instruction tuning, MoE models fall short in performance when compared to dense models on downstream tasks. 
        \item We highlight that when supplemented with instruction tuning, MoE models exceed the performance of dense models on downstream tasks, as well as on held-out zero-shot and few-shot tasks.
    \end{itemize}
    \item We present a comprehensive series of experiments, offering a comparative analysis of the performance of diverse MoE models subjected to instruction-tuning.
\end{itemize}
\section{Method}
\label{sec:method}

\subsection{Model Architecture} 
We leverage sparsely activated Mixture-of-Experts (MoE)~\cite{gshard,switchtransformer,expertchoice} in \shortname{}
models. 
Similar to the Switch Transformer~\cite{switchtransformer}, we replace the feed-forward component of every other Transformer layer with an MoE layer. Each MoE layer consists of a collection of
independent feed-forward networks as the `experts'. 
A gating function then uses a softmax activation function
to model a probability distribution over these experts.
This distribution indicates how well each expert is able to
process the incoming input. 
Even though each MoE layer has many more parameters,
the experts are sparsely activated. This means that for a
given input token, only a limited subset of experts is used,
giving the model more capacity while limiting computation. In our architecture, the subset size is either one or two depending on the routing strategy. 
Each MoE layer’s learnable gating network is trained to use its input
to activate the best two experts for each token of an input sequence. During inference, the learned gating network dynamically picks the two best experts for each token. For an MoE layer with $E$ experts, this essentially provides a collection of $O(E^2)$ different combinations of feed-forward networks instead of one in the classic Transformer architecture, enabling greater computational flexibility. 
The final learned representation of a token will be the weighted
combination of the outputs from the selected experts.

\begin{figure*}[t]
\centering
    \renewcommand\tabcolsep{1pt}
\begin{tabular}{cc}
\begin{subfigure}[b]{0.49\textwidth}
\includegraphics[width=\textwidth]{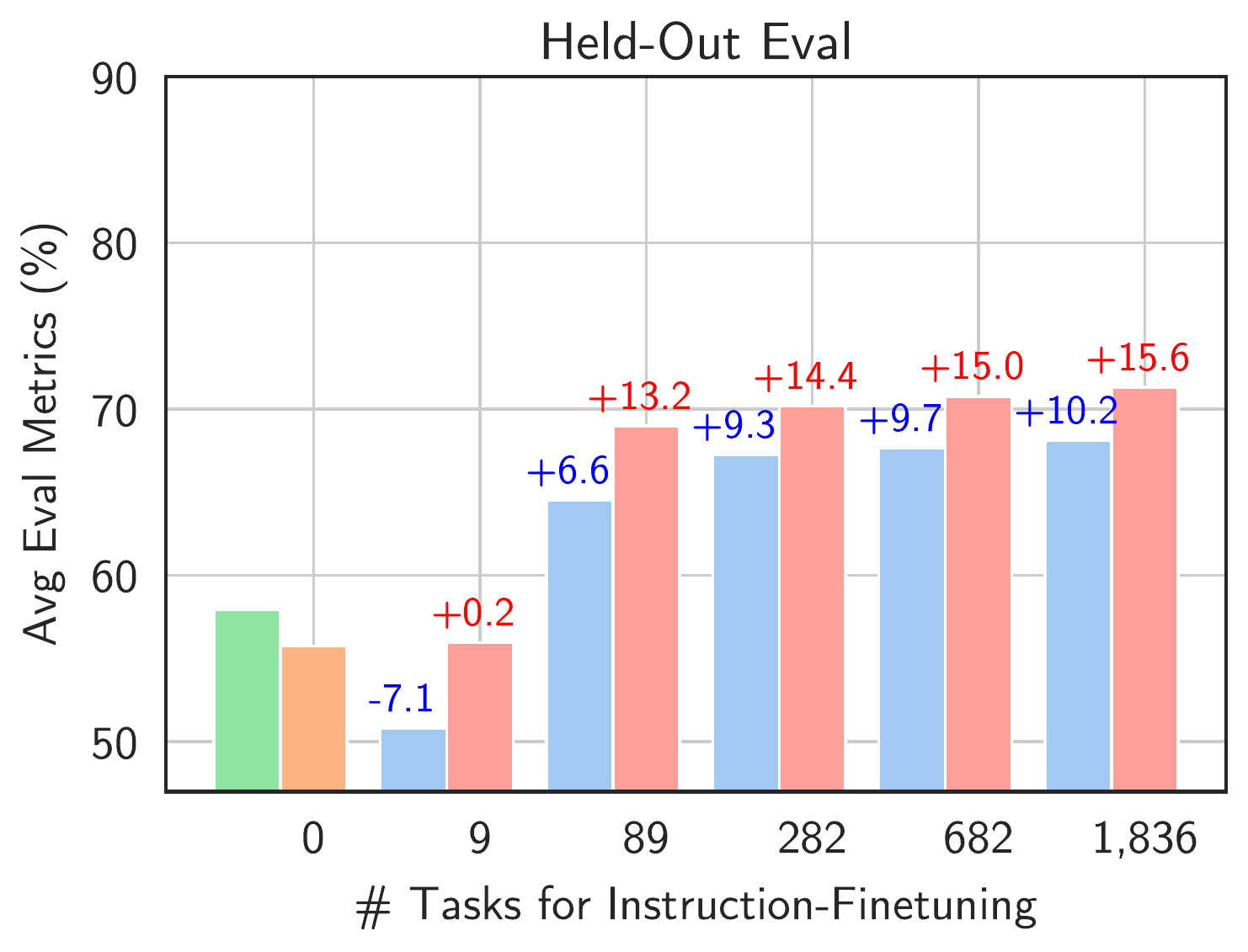}
\end{subfigure} 
&
\begin{subfigure}[b]{0.49\textwidth}
\includegraphics[width=\textwidth]{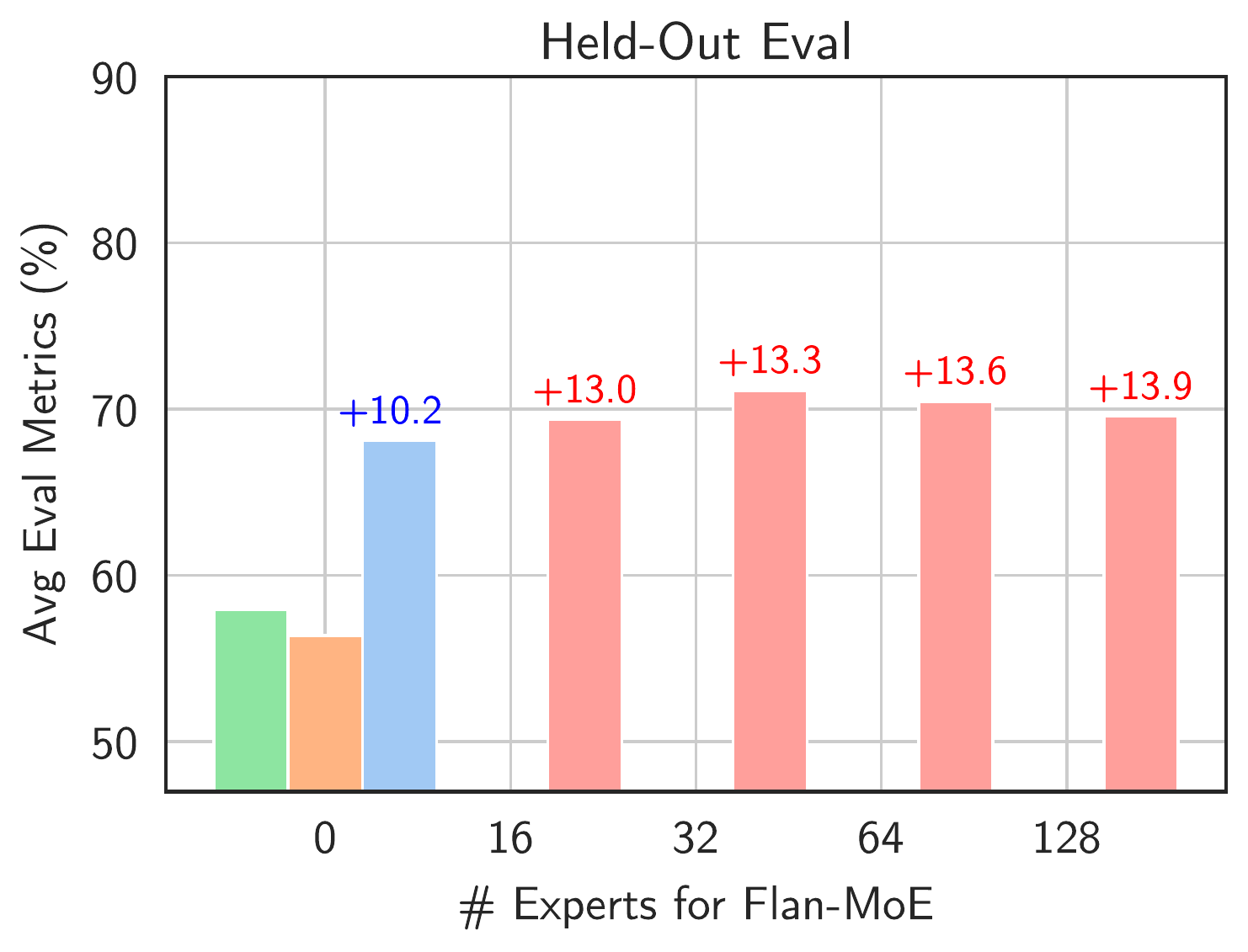}
\end{subfigure}
\\
& \hspace{-70mm} \includegraphics[width=0.8\textwidth]{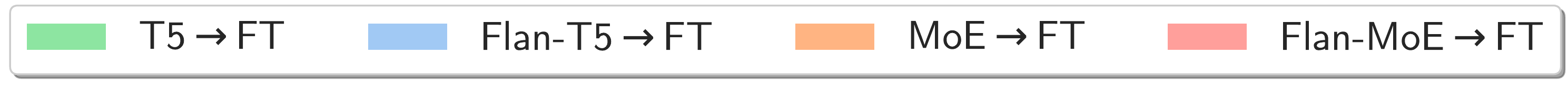} 
\end{tabular}
\caption{The effect of instruction tuning on \textsc{MoE} models versus dense counterparts for base-size models (same flops across all models in this figure). We perform single-task finetuning for each model on held-out benchmarks. \textbf{Compared to dense models, MoE models benefit more from instruction-tuning, and are more sensitive to the number of instruction-tuning tasks.} Overall, the performance of MoE models scales better with respect to the number of tasks, than the number of experts.}
\label{fig:main-finetune-gap-flan}
\end{figure*}

\subsection{Instruction Fine-tuning Recipe} 
We fine-tune \shortname{} using the prefix language model objective on the FLAN collective dataset~\cite{flant5,longpre2023flan}. 
Each \shortname{} will inherit the auxiliary loss setting during pre-training. 
All the model parameters will be updated. 
We adapt the sequence length of each \shortname{} to $2,048$ for input and $512$ for output based on the relative position embedding.
The dropout rate is $0.05$ and the expert dropout rate is $0.2$. 
The learning rate is $1e^{-4}$. 
The optimizer setting follows~\cite {flant5}.
\section{Experiment}
\label{sec:experiment}

\begin{table*}
\resizebox{\textwidth}{!}{%
\centering
\small
\centering
\begin{tabular}{@{}lcccccccccc}
\toprule
\multirow{2}{*}{\bf Model} & \bf FLOPs & \bf Total & \multicolumn{2}{c}{\bf MMLU} & \multicolumn{2}{c}{\bf BBH} & \bf Reasoning & \bf QA & \multirow{2}{*}{\bf Norm. Avg.} \\
 & \bf per token & \bf \# Params &  Direct &  CoT & Direct &  CoT & CoT & Direct \\
\midrule
T$5_\textsc{small}$ & 0.06G  & 80M & 26.7 & 7.2 & 26.7 & 5.6 & 10.3 & 33.8 & 26.3 \\\vspace{3mm} 
\sshortname{}-T$5_\textsc{small}$ & 0.06G  & 80M & 28.7 & 12.1 & 29.1 & 19.2 & 15.0 & 40.9 & 28.7 \bluegain{2.4} \\
T$5_\textsc{base}$ & 0.3G & 250M & 25.7 & 14.1 & 27.7 & 14.6 & 14.7 & 35.3 & 26.2 \\\vspace{3mm} 
\sshortname{}-T$5_\textsc{base}$ & 0.3G & 250M & 35.6 & 33.3 & 30.3 & 26.8 & 16.4 & 48.8 & 33.9 \bluegain{7.7} \\
T$5_\textsc{large}$ & 1.0G & 780M & 25.1 & 15.3 & 27.7 & 16.2 & 11.9 & 36.4 & 25.7 \\\vspace{3mm} 
\sshortname{}-T$5_\textsc{large}$ & 1.0G & 780M & 44.7 & 38.9 & 34.7 & 28.5 & 22.2 & 64.6 & 42.0 \bluegain{16.3} \\
T$5_\textsc{xl}$ & 3.6G & 3B & 25.3 & 14.1 & 27.4 & 19.3 &  14.2 & 38.2 & 25.9 \\\vspace{3mm} 
\sshortname{}-T$5_\textsc{xl}$ & 3.6G & 3B & 50.3 & 46.1 & 40.2 & 35.9 & 33.9 & 74.1 & 48.0 \bluegain{22.1} \\
T$5_\textsc{xxl}$ & 13.9G & 11B & 26.1 & 19.1 & 29.5 & 19.3 & 21.4 & 47.4 &  27.7 \\\vspace{3mm} 
\sshortname{}-T$5_\textsc{xxl}$ & 13.9G & 11B & 52.6 & 47.9 & 45.6 & 41.6 & 46.3 & 80.4 & 51.7 \bluegain{24.0}  \\
PaLM& 12.6G & 8B&  24.3 & 24.1 & 30.8 &  30.1 & 24.9 & 47.6 & 27.1 \\\vspace{3mm} 
\sshortname{}-PaLM& 12.6G & 8B& 49.3& 41.3& 36.4& 31.1& 36.9  & 75.1 & 47.5 \bluegain{20.4} \\
PaLM& 91.6G & 62B& 55.1 & 49.0 & 37.4 & 43.0 & 50.6 & 70.4 & 51.0 \\\vspace{3mm} 
\sshortname{}-PaLM& 91.6G & 62B& 59.6& 56.9& 47.5& 44.9& 59.7  & 85.3 & 57.6 \bluegain{6.6} \\
PaLM& 847G & 540B& 71.3 & 62.9 & 49.1 & 63.7 & 72.6 & 86.0 & 66.2 \\
\sshortname{}-PaLM& 847G & 540B& 73.5& 70.9& 57.9& 66.3& 76.5  & 89.9 & 70.3 \bluegain{4.1} \\
\midrule 
Switch$_\textsc{base}$ & 0.3G & 3.5B & 28.3 & 13.6 & 0.1 & 1.4 & 5.2 & 35.8 & 20.2 \\\vspace{3mm}
\sshortname{}-Switch$_\textsc{base}$ & 0.3G & 3.5B & 38.0 & 34.2 & 33.2 & 29.4 & 18.6 & 58.0 & 36.8 \bluegain{16.6} \\
Switch$_\textsc{large}$ & 1.0G & 26B & 24.0 & 23.1 & 0.2 & 7.2 & 12.4 & 33.7 & 17.7 \\\vspace{3mm}
\sshortname{}-Switch$_\textsc{large}$ & 1.0G & 26B & 46.1 & 40.3 & 36.3 & 28.0 & 25.3 & 66.5 & 43.5 \bluegain{25.8} \\
Switch$_\textsc{xxl}$ & 13.9G & 395B & 24.6 & 15.1 & 0.0 & 6.7 & 9.2 & 32.5 & 17.8 \\
\sshortname{}-Switch$_\textsc{xxl}$ & 13.9G & 395B & 55.6 & 50.1 & 47.9 & 43.5 & 46.6 & 78.8 & 54.2 \bluegain{36.4} \\
\midrule
GS$_\textsc{small}$ & 0.06G & 0.3B & 23.9 & 0.0 & 0.2 & 0.8 & 0.8 & 24.1 & 16.7 \\\vspace{3mm}
\sshortname{}-GS$_\textsc{small}$ & 0.06G & 0.3B & 32.6 & 26.9 & 29.6 & 20.9 & 16.1 & 48.9 & 31.8 \bluegain{15.1} \\
GS$_\textsc{base}$ & 0.3G & 1.3B & 25.0 & 15.9 & 0.0 & 4.8 & 3.8 & 26.8 & 17.6 \\\vspace{3mm}
\sshortname{}-GS$_\textsc{base}$ & 0.3G & 1.3B & 39.9 & 33.6 & 33.7 & 25.1 & 22.0 & 57.9 & 38.3 \bluegain{20.7} \\
GS$_\textsc{large}$ & 1.0G & 9.2B & 26.4 & 12.8 & 0.2 & 14.3 & 13.0 & 31.9 & 19.2 \\\vspace{3mm}
\sshortname{}-GS$_\textsc{large}$ & 1.0G & 9.2B & 47.8 & 40.8 & 35.0 & 29.2 & 27.6 & 69.5 & 44.5 \bluegain{25.3} \\
GS$_\textsc{xl}$ & 03.6G & 17.4B & 25.7 & 10.0 & 0.0 & 0.0 & 10.4 & 35.0 & 18.7 \\
\sshortname{}-GS$_\textsc{xl}$ & 3.6G & 17.4B & 51.1 & 42.3 & 40.1 & 31.4  & 34.3 & 73.9 & 48.7 \bluegain{30.0}\\
\midrule
EC$_\textsc{small}$ & 0.06G & 0.3B & 25.3 & 1.2 & 0.1 & 2.3 & 0.8 & 36.0 & 18.1  \\\vspace{3mm}
\sshortname{}-EC$_\textsc{small}$ & 0.06G & 0.3B & 34.1 & 25.1 & 29.2 & 22.1 & 16.6 & 58.1 & 33.1 \bluegain{15.0}  \\
EC$_\textsc{base}$ & 0.3G & 1.3B & 25.0 & 25.9 & 0.0 & 1.4 & 14.3 & 35.7 & 18.5 \\\vspace{3mm}
\sshortname{}-EC$_\textsc{base}$ & 0.3G & 1.3B & 42.7 & 33.0 & 34.0 & 26.7 & 22.2 & 61.5 & 40.3 \bluegain{21.8} \\
EC$_\textsc{large}$ & 1.0G & 9.2B & 23.4 & 12.6 & 0.0 & 8.6 & 6.7 & 40.1 & 17.3 \\\vspace{3mm}
\sshortname{}-EC$_\textsc{large}$ & 1.0G & 9.2B & 48.3 & 44.5 & 37.9 & 32.0 & 32.2 & 73.1 & 46.4 \bluegain{29.1} \\
EC$_\textsc{xl}$ & 3.6G & 17.4B & 26.7 & 11.0 & 0.0 & 1.9 & 12.4 & 34.2 & 19.4 \\
\sshortname{}-EC$_\textsc{xl}$ & 3.6G & 17.4B & 52.1 & 41.4 & 40.3 & 33.2 & 38.1 & 74.3 & 49.4 \bluegain{30.0}\\
\midrule
ST$_\textsc{base}$ & 0.3G & 1.3B & 25.2 & 17.7 & 0.0 & 14.0 & 12.6 & 25.7 & 18.1 \\\vspace{3mm}
\sshortname{}-ST$_\textsc{base}$ & 0.3G & 1.3B & 42.4 & 35.5 & 34.9 & 26.4 & 22.5 & 61.5 & 40.4 \bluegain{21.8} \\
ST$_\textsc{32B}$ & 32.1G & 259B & 25.5 & 15.1 & 0.0 & 5.5 & 9.8 & 32.1 & 18.4  \\
\sshortname{}-ST$_\textsc{32B}$ & 32.1G & 259B & 65.4 & 63.0 & 54.4 & 47.4 & 66.3 & 63.9 & 63.6 \bluegain{45.2} \\
\bottomrule
\end{tabular}}
\caption{
MoE models improve instruct fine-tuning performance on top of dense counterparts. 
The benchmark suites are MMLU (57 tasks), BBH (23 tasks), Reasoning (4 Tasks), and QA (4 Tasks). 
The evaluation metric across all benchmarks is few-shot prompted accuracy, specifically the exact match. 
To calculate this metric, we take an unweighted average across all tasks. For a comprehensive evaluation, we report the normalized average of MMLU-direct, BBH-direct, Reasoning-CoT, and QA-Direct. 
The MMLU and BBH evaluation benchmarks are held-out (not included in the finetuning data.) while the Reasoning and QA evaluation benchmarks are held-in. 
(Noted that \sshortname{}-ST$_\textsc{32B}$ outperforms \flanpalm$_\textsc{62B}$ while being <30\% of the FLOPS.)
}
\label{tbl:results:moe_main_task}
\end{table*}

We study \shortname{} in the context of instruction-tuning. 
We first perform a controlled comparison of \shortname{} to an equivalent “standard” dense encoder-decoder Transformer (T5), across a range of model sizes in Section~\ref{subsec:control_study}.
We subsequently demonstrate in Section~\ref{subsec:scale_up} that scaling up our model, referred to as \shortname{}, can attain remarkable performance levels. Our most extensive model, \sshortname{}-ST$_\textsc{32B}$, surpasses the performance of \flanpalm$_\textsc{62B}$ while utilizing less than 30\% of FLOPs per token.
We further ablate the various design decisions in the next Section.

\subsection{Settings}

\paragraph{Traning Data.} By default, all models are trained on the 1,836 finetuning tasks by combining four mixtures from prior work: Muffin, T0-SF, NIV2, and CoT, as  in \cite{flant5}. Specifically, Muffin comprises 80 tasks from \cite{flan} and 26 dialog/program synthesis tasks; T0-SF comprises 193 tasks from \cite{t0}; NIV2 comprises 1554 tasks from \cite{naturalinstructions}; CoT comprises 9 reasoning tasks. 

\paragraph{Evaluations.} We conduct both zero-shot and few-shot evaluations on held-out tasks as in \cite{flant5} which were not included as part of the finetuning data. We use MMLU \cite{mmlu} that includes exam questions from 57 tasks such as mathematics, history, law, and medicine; 
BBH includes 23 challenging tasks from BIG-Bench \cite{bigbench}; 
The reasoning benchmark comprises four tasks: GSM8K~\cite{gsm8k} and SVAMP~\cite{svamp}/ASDIV~\cite{asdiv} incorporate the grade school math word problems and the elementary-level math word problems, and StrategyQA~\cite{strategyqa} measures open-domain questions where the required reasoning steps are implicit in the question;
The QA benchmark include four QA tasks: the elementary AI2 science category in UnifiedQA~\cite{unifiedqa},  BoolQ~\cite{boolq}, ARC-easy and ARC-challenge~\cite{arc} that covers QA tasks in abstract, yes/no, multiple-choice formats.   
For MMLU and BBH, we evaluate both the ability of directly predicting the answer via direct prompting, where the model directly gives the answer~\cite{flant5}, as well as via chain-of-thought (CoT) prompting, where the model must provide a reasoning chain before giving the final answer~\cite{wei2022chain}. For reasoning tasks, we only measure CoT prompting accuracy. 
For all benchmarks except for QA we use the given few-shot exemplars, with the number of exemplars
following prior work: five-shot for MMLU, three-shot for BBH, eight-shot for reasoning tasks, and zero-shot for QA. 
For a given model we also report a single “normalized average” metric, following the “normalized preferred metric” in BIG-Bench \cite{bigbench}. 
Our normalized average metric is the macro-average over four normalized scores: MMLU-Direct, BBH-Direct, Reasoning-CoT, and QA-Direct.
Results for all tasks in each benchmark are reported in Appendix. 

\subsection{Controlled study across scales}
\label{subsec:control_study}
\begin{figure*}[tb]
\centering
    \renewcommand\tabcolsep{1pt}
\begin{tabular}{cc} 

& \hspace{-70mm} \includegraphics[width=\textwidth]{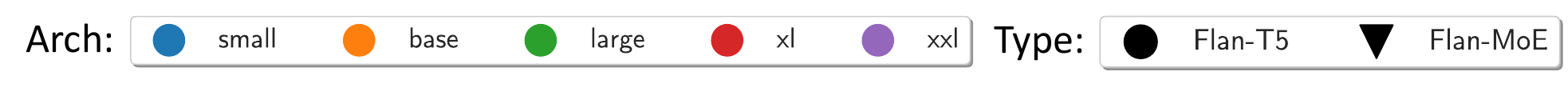} \\
\begin{subfigure}[b]{0.45\textwidth}
\includegraphics[width=\textwidth]{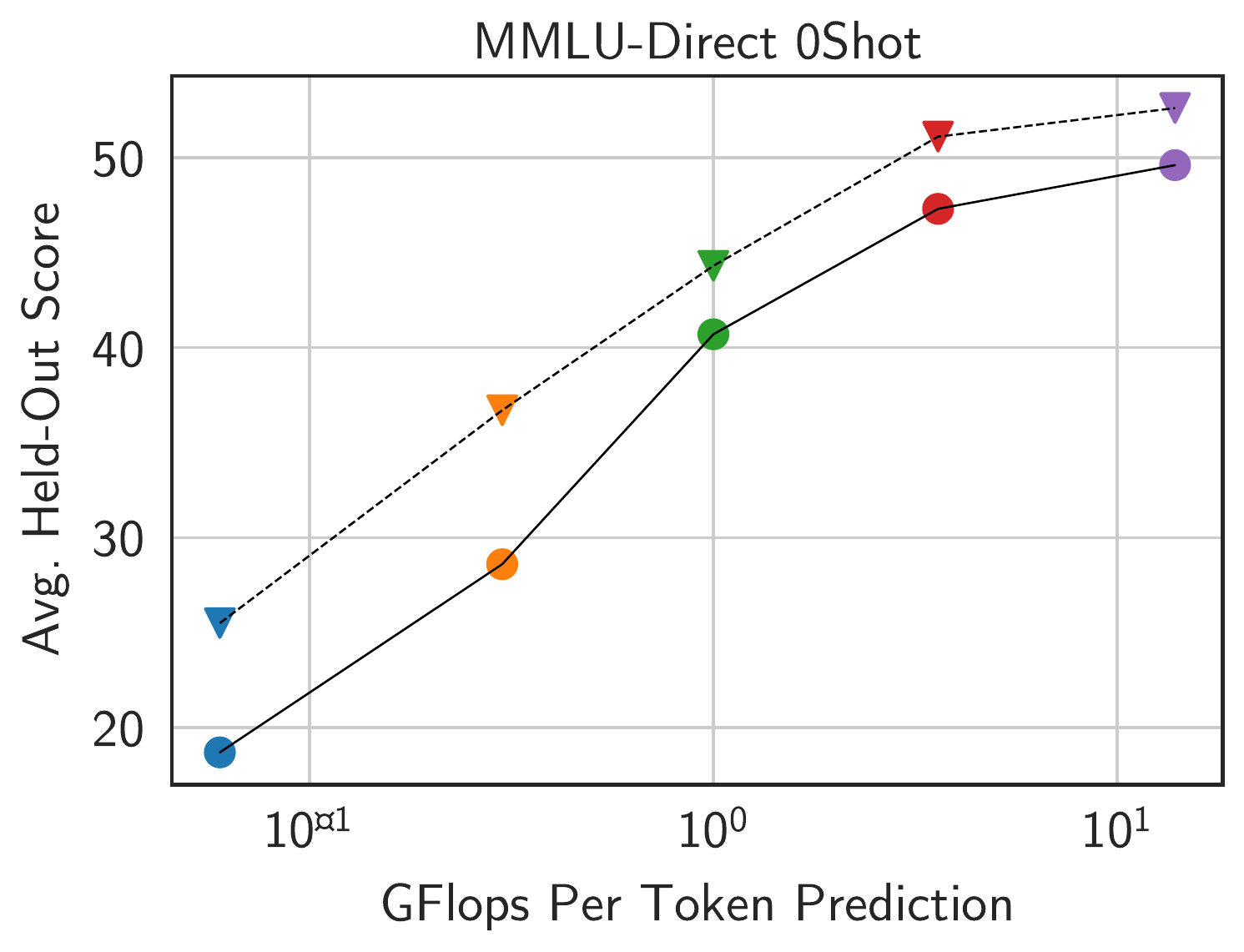}
\end{subfigure}
&
\begin{subfigure}[b]{0.45\textwidth}
\includegraphics[width=\textwidth]{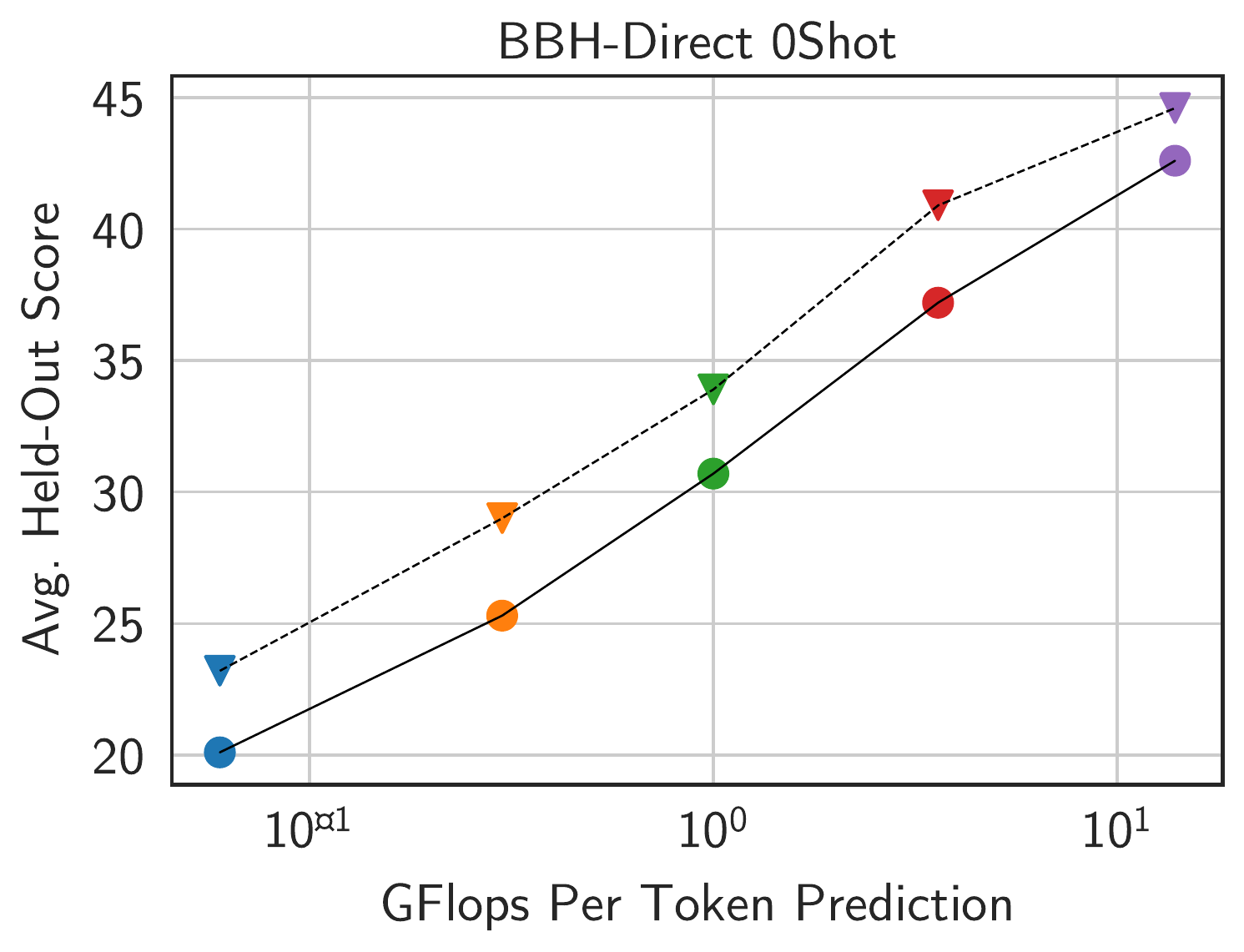}
\end{subfigure} 
\end{tabular}
\caption{Average zero performance of \shortname{} models versus \textsc{Flan}-T5 dense models for similar effective FLOPs per token over the 57 MMLU tasks and 23 BBH tasks.\protect\footnotemark}
\label{fig:scale-size-data-efficiency}
\end{figure*}

We instruction finetune a range of \shortname{} models at batch size 32 and sequence length 2048 for 200k steps. 
This matches the number of training examples used for \textsc{Flan}-T5~\cite{flant5}. 
We re-finetuning our own \textsc{Flan}-T5 variants for fair comparisons. 
\footnotetext{We use 64 experts for \textsc{small}, \textsc{base}, \textsc{32b}, \textsc{xl} and 128 experts for all the other model sizes following~\cite{switchtransformer,expertchoice,stmoe}}

\paragraph{Dense Model Size.} 
Figure~\ref{fig:scale-size-data-efficiency} shows the performance of each model (dense and sparse) against forward-pass FLOPs.
The cost-performance Pareto frontier for \shortname{} dominates the dense models by a wide margin, indicating that \shortname{} offers strong improvements across all scales from small, up to xxl. 
The effect is particularly large on zero-shot
and few-shot MMLU-Direct, with absolute performance improvements of 7.1\% 
on average. 
For challenging tasks in BBH-Direct, \shortname{} offers a strong boost at small scales, while
at larger scales the gains are more modest but still significant.

\begin{figure*}[tb]
\centering
    \renewcommand\tabcolsep{1pt}
\begin{tabular}{cccc} 
\begin{subfigure}[b]{0.25\textwidth}
\includegraphics[width=\textwidth]{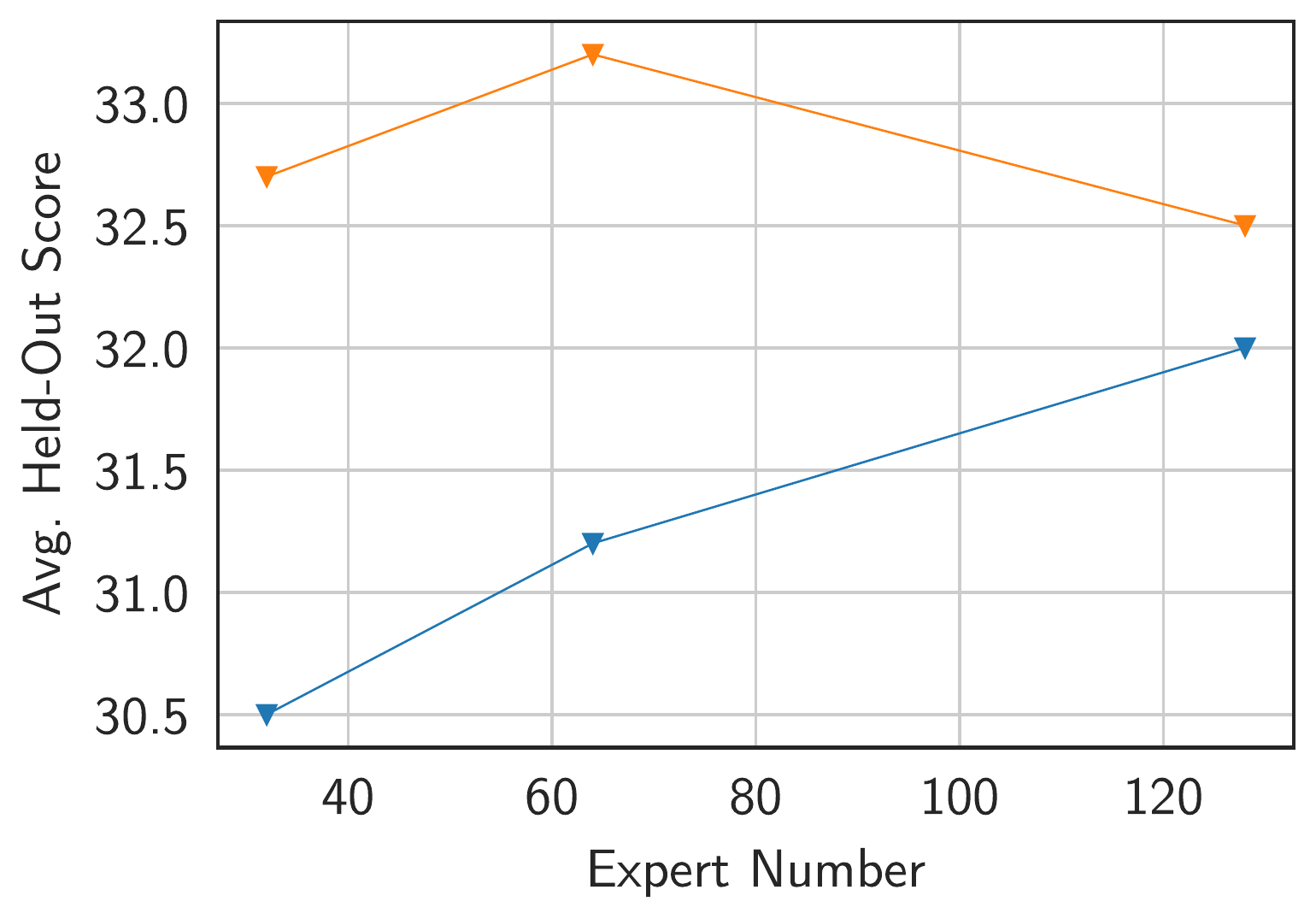}
\caption{Scaling (MMLU)}
\end{subfigure}
&
\begin{subfigure}[b]{0.25\textwidth}
\includegraphics[width=\textwidth]{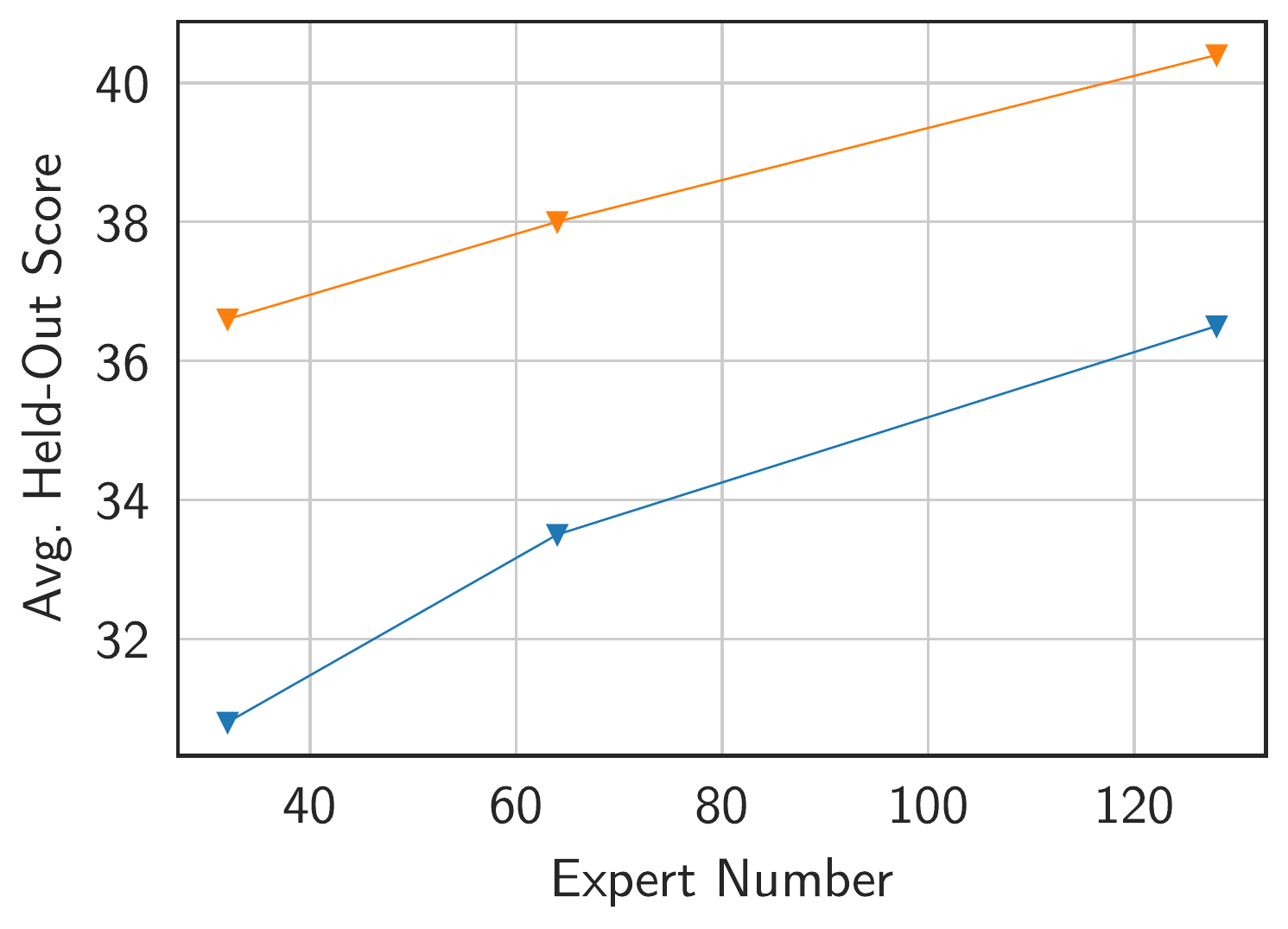}
\caption{Scaling (BBH)}
\end{subfigure} 
&
\begin{subfigure}[b]{0.25\textwidth}
\includegraphics[width=\textwidth]{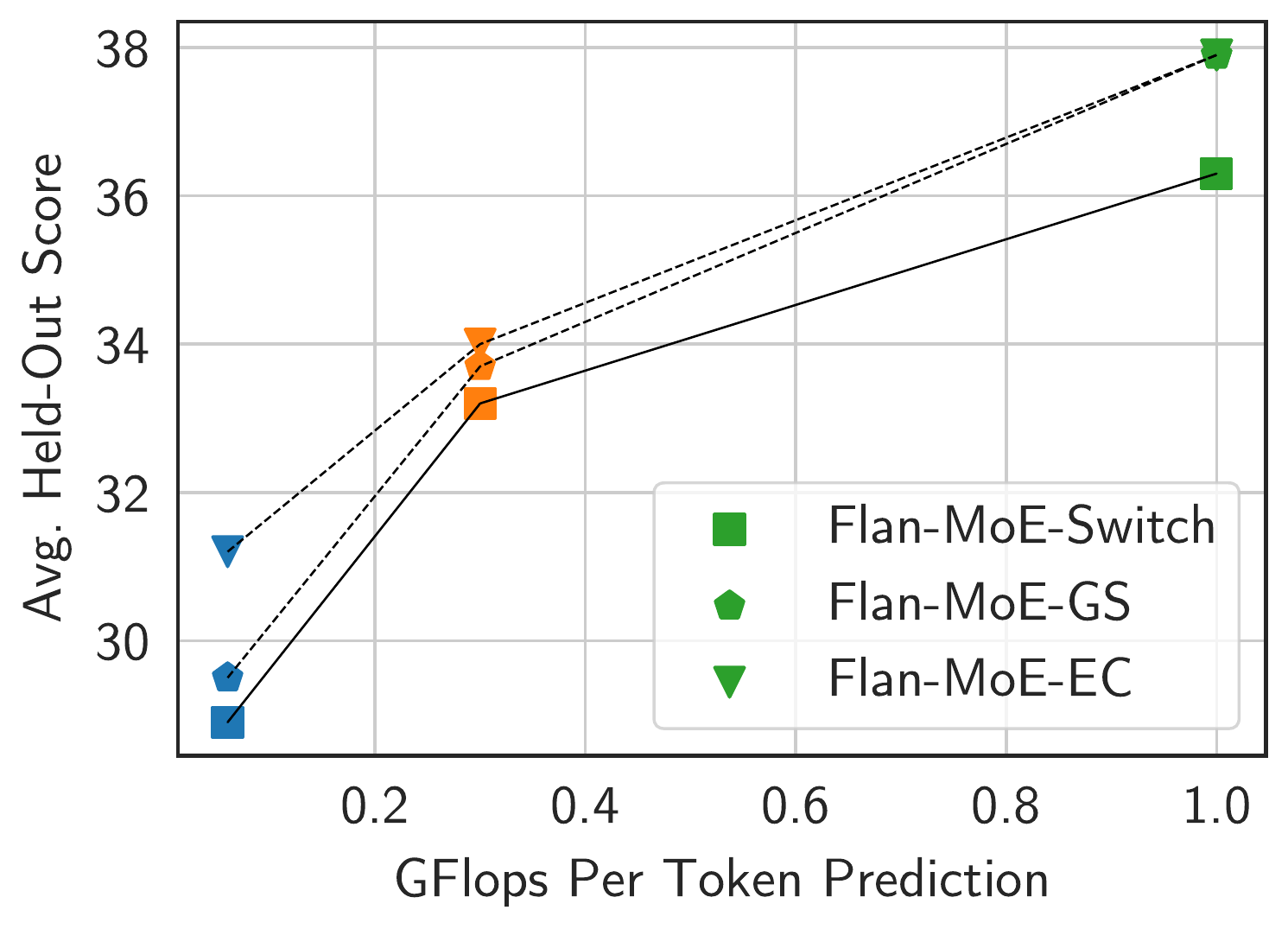}
\caption{Routing (MMLU)}
\end{subfigure}
&
\begin{subfigure}[b]{0.25\textwidth}
\includegraphics[width=\textwidth]{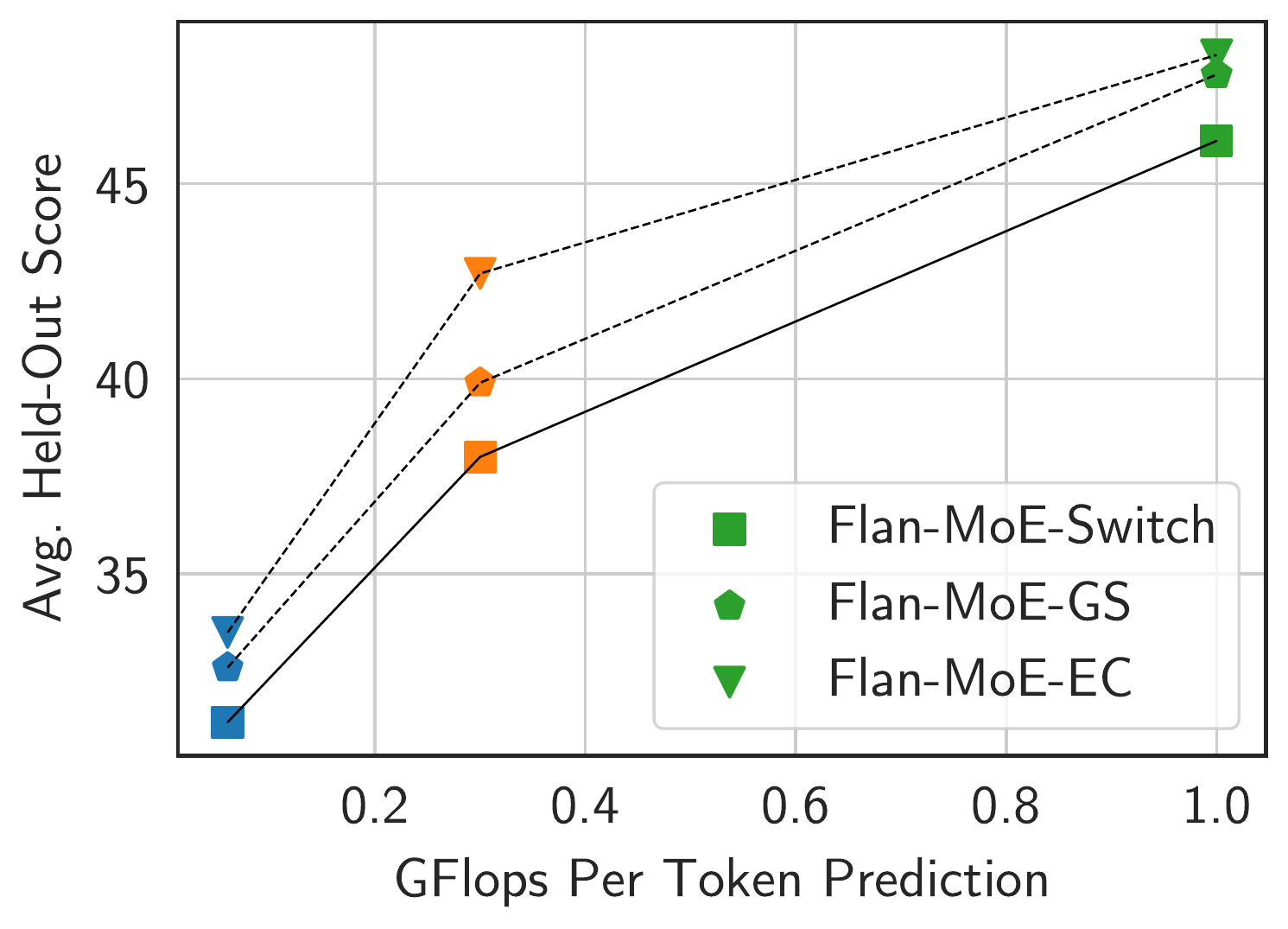}
\caption{Routing (BBH)}
\end{subfigure} 
\end{tabular}
\caption{Average few-shot performance of \shortname{} models over the 57 MMLU tasks and 23 BBH tasks. (Different color represents different dense model sizes.)}
\label{fig:expert-routing-scale}
\end{figure*}

\paragraph{Expert Number.}  
The performance of \shortname{} models has been observed to scale with the number of experts included in the architecture, but it tends to saturate beyond a certain threshold. 
Initially, as the number of experts increases in Figure~\ref{fig:expert-routing-scale}, the model benefits from a richer repertoire of specialized sub-networks, each capable of handling distinct tasks or aspects of the problem space. This diverse ensemble enables the MoE model to demonstrate enhanced adaptability and efficiency in processing complex tasks, leading to improved performance overall. However, as the number of experts continues to grow, the performance gains begin to diminish, eventually reaching a point of saturation for \textsc{base}-sized model. 

\paragraph{Routing Strategy}

Routing strategy is an essential component of Mixture-of-Experts (MoE) models, playing a pivotal role in determining the effectiveness and efficiency of these models. 
The primary function of the routing strategy is to intelligently distribute input data among multiple specialized experts, each optimized for handling specific subsets of the input space. 
This distribution process is crucial for maximizing the utilization of the model's capacity while minimizing the risk of overfitting. 
An effective routing strategy not only ensures that the appropriate experts are selected for a given input, but also that resources are allocated optimally, leading to enhanced computational efficiency and faster training times. 
Consequently, there have been two trending strategies, token-choice~\cite{gshard} which lets the token select the top-$K$ experts, and expert-choice~\cite{expertchoice} which lets the experts select the top-$K$ tokens. 

\begin{figure*}[tb]
\centering
    \renewcommand\tabcolsep{1pt}
\begin{tabular}{cc} 
\begin{subfigure}[b]{0.45\textwidth}
\includegraphics[width=\textwidth]{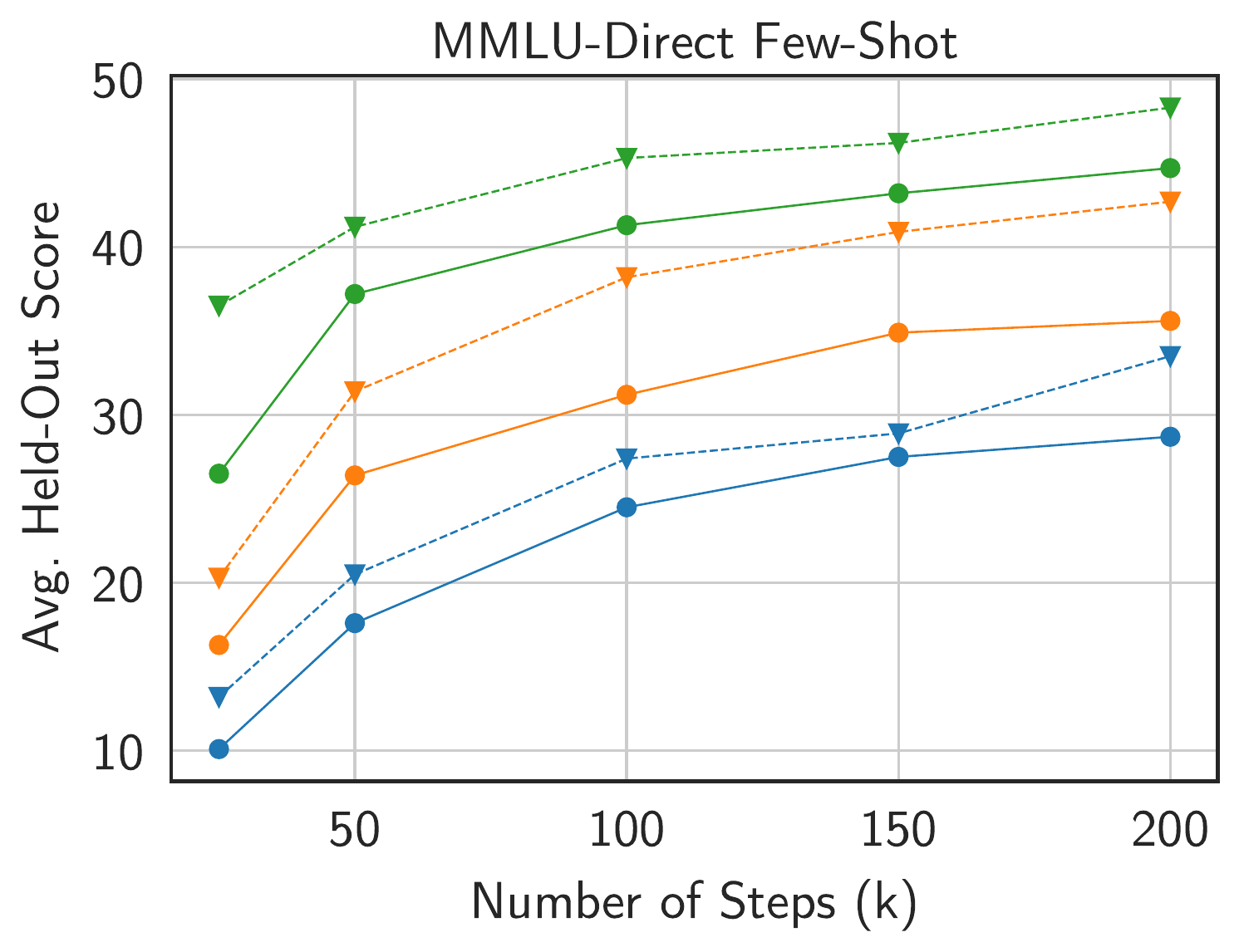}
\end{subfigure}
&
\begin{subfigure}[b]{0.45\textwidth}
\includegraphics[width=\textwidth]{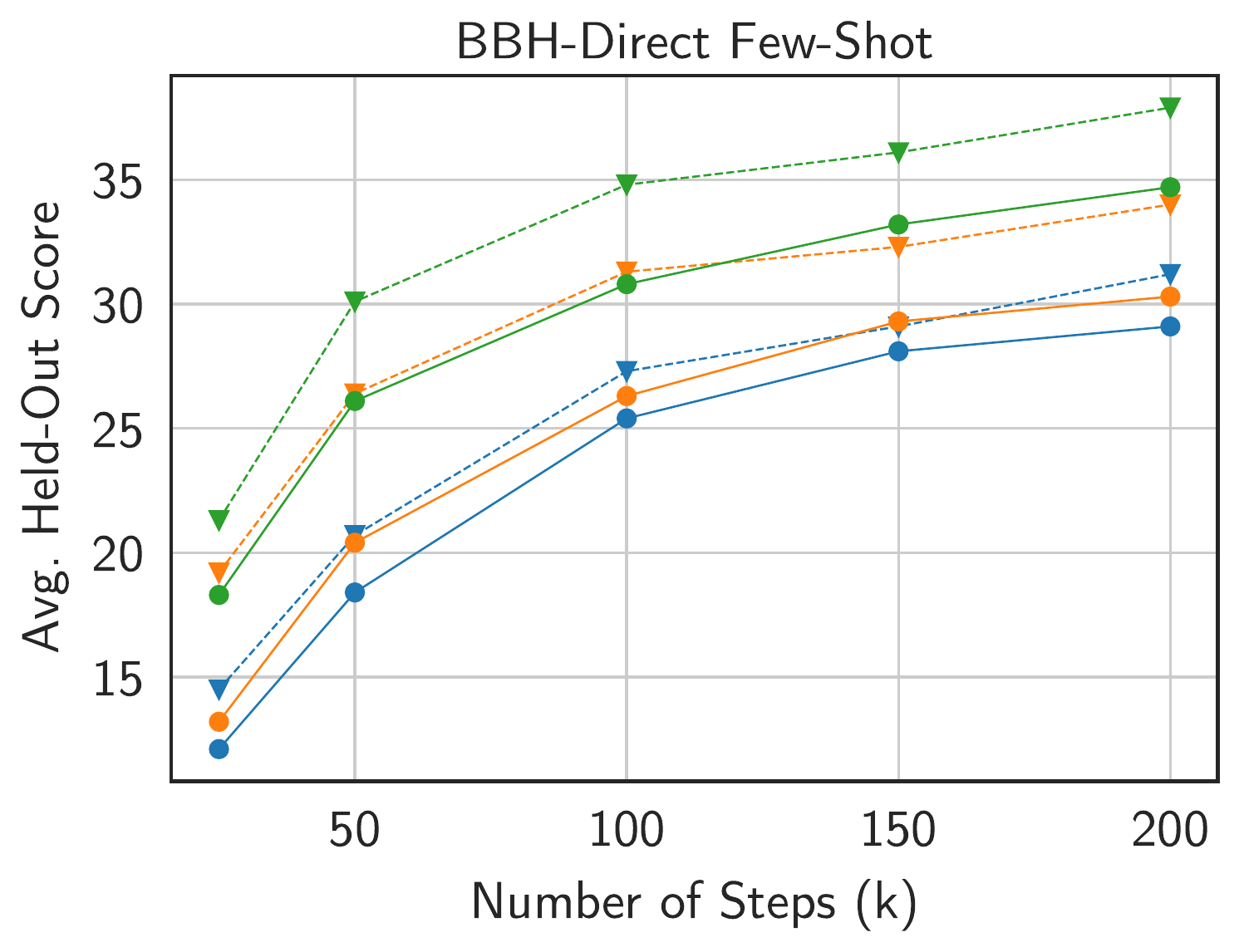}
\end{subfigure} 
\end{tabular}
\caption{\label{figs:efficiency} Learning efficiency comparison. Average zero-shot, and few-shot performance of \shortname{} models versus \textsc{Flan}-T5 dense models as more tokens are processed during training on FLAN Tasks. }
\end{figure*}

We presented a detailed study about how different routing decisions affect the instruct fine-tuning performance in Figure~\ref{figs:efficiency} and Table~\ref{tbl:results:moe_main_task}, which includes the checkpoints from Switch Transformer top-1 token-choice gating (\textsc{Flan}-Switch), GShard top-2 token-choice gating (\textsc{Flan}-GS) and expert-choice top-2 gating (\textsc{Flan}-EC) models pre-trained on the same GLaM~\cite{glam} dataset. 
It is evident that activating more experts, as demonstrated by the comparison between the \textsc{Flan}-Switch and \textsc{Flan}-GS strategies, results in enhanced performance across all four benchmarks. Among these benchmarks, the MMLU-Direct model shows the most significant improvement, with an increase from 38.0\% to 39.9\% for \textsc{base/large}-sized models. Although the gains at the extra-large scale are more modest, they remain noteworthy and meaningful.
It's noteworthy that instruction-tuning significantly amplifies the performance of both held-out MMLU, BBH, and held-in QA and reasoning benchmarks for MoE models in comparison to dense models of equivalent capacity. The advantages are amplified even further for larger MoE models. For instance, instruction-tuning enhances the performance of $\textsc{ST}_\textsc{32B}$ by a substantial 45.2\%, while the improvement observed for \flanpalm$_\textsc{62B}$ is comparatively modest at around 6.6\%.

Furthermore, the \textsc{Flan}-EC strategy consistently outshines the \textsc{Flan}-GS approach for the given model across various scales and tasks. 
It is noteworthy that the performance gap between the token-choice and expert-choice models can be bridged when we incorporate advanced auxiliary loss and pre-training strategy as exhibited in \textsc{ST-MoE}~\cite{stmoe}. 
This integration led to the development of our \textsc{Flan}-ST models. Considering that the largest \textsc{ST-MoE} set the benchmark in a variety of NLP tasks when appropriately fine-tuned, we have also decided to scale up \textsc{Flan}-ST, employing instruction fine-tuning.

\subsection{Scaling up \shortname{}}
\label{subsec:scale_up}
We increase the architecture size to assess the performance of \shortname{} in the large-scale regime. 
As discussed above, we instruction fine-tune the largest ST-MoE$_\textsc{32B}$~\cite{stmoe} model with 12 expert layers in encoder, and decoder, respectively; these are non-uniformly distributed, with 64 experts per layer, and $K=2$ activated per token. 
It was trained at a batch size of 32 and sequence length of 2048 for 200k steps. We average checkpoints towards the end of training. 
The model \textsc{Flan}-ST$_\textsc{32B}$, comprising a total of 32 billion parameters, only utilizes 32.1 GFLOPs per token, which amounts to merely one-third of the computational power required by a \flanpalm$_\textsc{62B}$ model. Additionally, all the routers combined account for less than 4 million parameters. Table~\ref{tbl:results:moe_main_task} illustrates the performance of this model alongside current state-of-the-art instruct fine-tuned models.

\textsc{Flan}-ST$_\textsc{32B}$ achieves a 65.4\% few-shot MMLU benchmark accuracy and a 54.4\% few-shot BBH benchmark accuracy, with a relatively modest architectural size and training count. 
Notably, \textsc{Flan}-ST$_\textsc{32B}$ surpasses the performance of \flanpalm$_\textsc{62B}$, which consumes nearly triple the compute resources, by a substantial margin across all four benchmarks. However, it is important to acknowledge the considerable performance gap that persists between the largest \flanpalm$_\textsc{540B}$ and \textsc{Flan}-ST$_\textsc{32B}$ models.

\section{Discussion}
\label{sec:discussion}

\subsection{Finetuing Strategy} 
Sparse models have performed remarkably well in the regime of large datasets, but have sometimes performed poorly when finetuning data is limited~\cite{stmoe,switchtransformer}. 
Instruction finetuning can also be viewed as a continual finetuning stage, so we present a detailed study about how different factors impact the instruct finetuning performance of \shortname{} and offer a practical recipe. 
All the discussion here is based on instruction finetuning \textsc{Flan}-EC$_\textsc{base}$/\textsc{Flan}-ST$_\textsc{base}$ for 100k steps. 

\begin{figure*}[tb]
\centering
    \renewcommand\tabcolsep{1pt}
\begin{tabular}{cc} 
\begin{subfigure}[b]{0.45\textwidth}
\includegraphics[width=\textwidth]{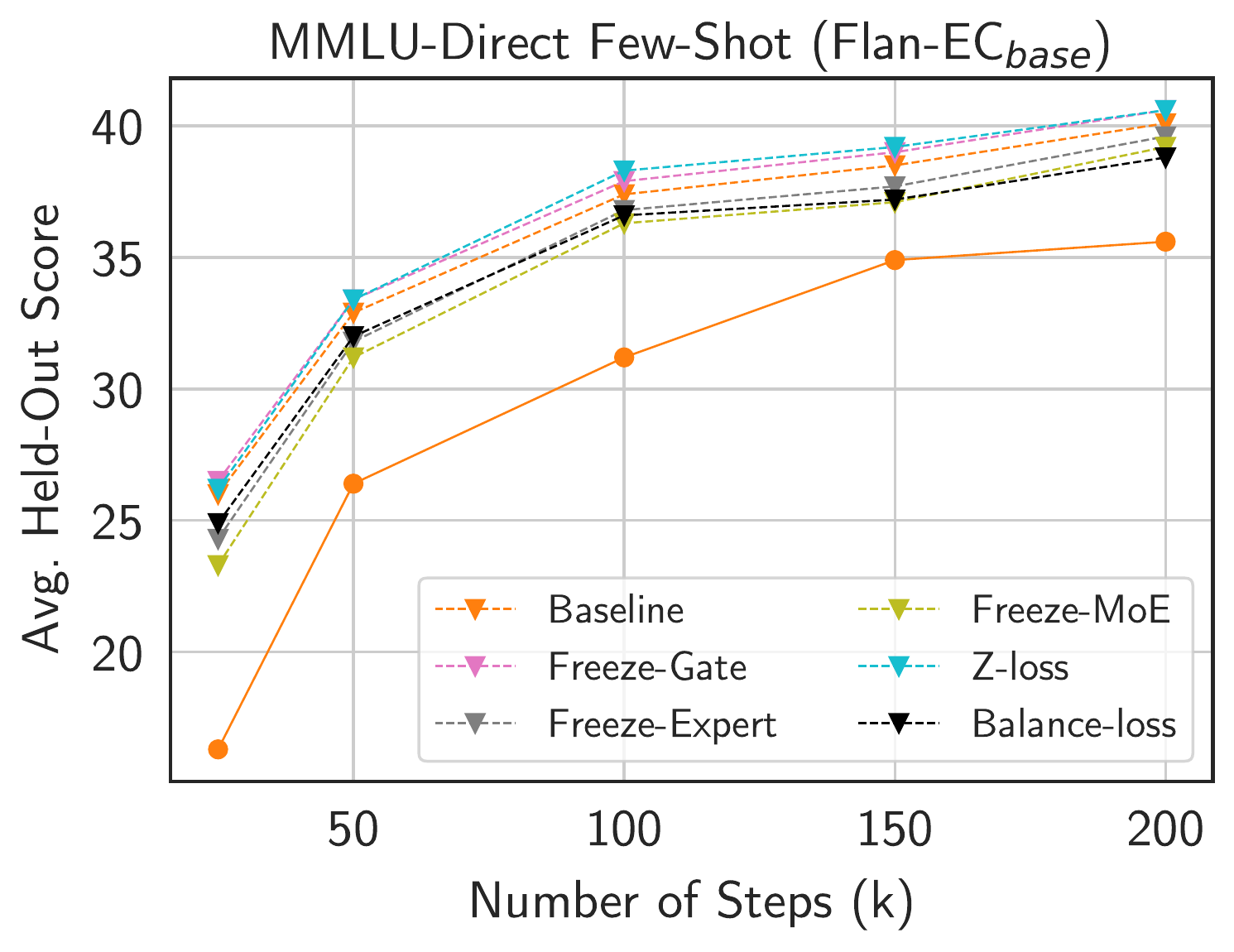}
\end{subfigure}
&
\begin{subfigure}[b]{0.45\textwidth}
\includegraphics[width=\textwidth]{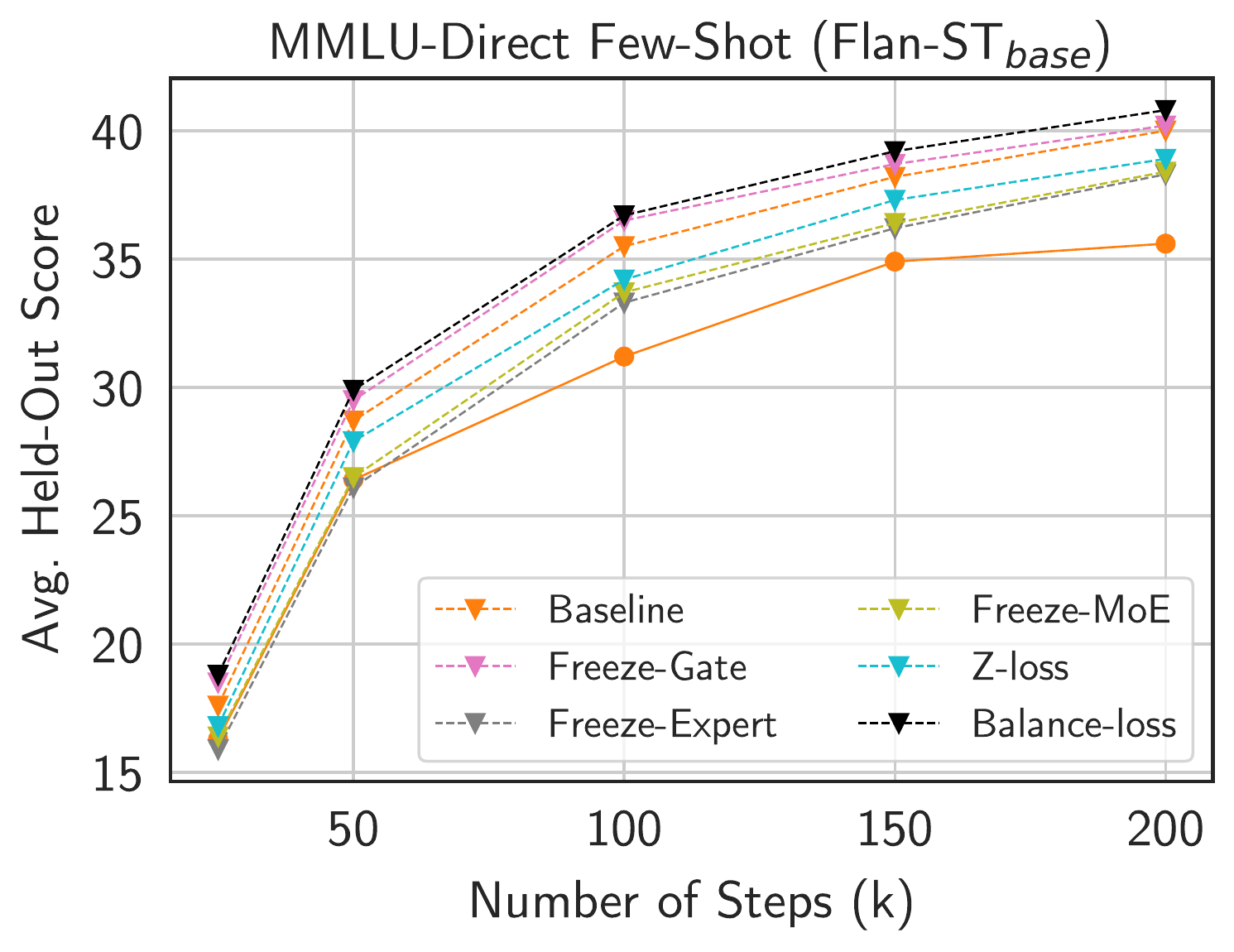}
\end{subfigure}
\end{tabular}
\caption{Average few-shot performance of \shortname{} with different finetuning strategy.}
\label{fig:ablation-finetune-data-efficiency}
\end{figure*}

\paragraph{Auxiliary Loss.}  The incorporation of auxiliary loss~\cite{gshard,stmoe} helps mitigate the risk of overfitting by promoting the diversification of the experts' knowledge and improving the model's generalization capabilities for sparsely gated mixture-of-expert models. Furthermore, auxiliary losses can be employed to address specific issues, such as load balancing among experts or preventing expert collapse, which can further enhance the model's overall performance. 
We experiment with both balancing loss that is used in~\cite{gshard} and router Z-loss that is used in~\cite{stmoe} in Table~\ref{tbl:ablation_freeze}. 
The implementation of balancing loss contributed to enhanced performance on MMLU, BBH, and GSM8K for \textsc{Flan}-EC$\textsc{base}$, whereas Z-loss resulted in a deterioration of performance. Conversely, for \textsc{Flan}-ST$\textsc{base}$, we observed a contrasting trend. 
We conjecture that the discordance between the auxiliary loss during pre-training and instruction-tuning could potentially disrupt the optimization process, thereby leading to a suboptimally optimized \shortname{} model. 

\paragraph{Expert/Gating Freeze.} 
In an effort to enhance the generalization capabilities of sparse models and combat overfitting, researchers have discovered that finetuning a subset of model parameters results in improved generalization performance for ST-MoE models, as noted in the study by ST-MoE~\cite{stmoe}. Interestingly, it was observed that updating non-MoE parameters yields similar outcomes to updating all parameters, while updating only expert parameters performs slightly better.

We conducted experiments by freezing the gating function, expert modules, and MoE parameters of the given model, as presented in Table~\ref{tbl:ablation_freeze}. The results indicate that freezing either the expert or MoE components negatively impacts performance. 
Conversely, freezing the gate slightly improves performance, albeit not significantly. 
We postulate that this observation is related to the under-fitting of the \shortname{}, as in Figure~\ref{fig:ablation-finetune-data-efficiency}, which depicts the finetuning data efficiency ablation study.

\begin{table}[t]
\centering
\begin{tabular}{cc}
\resizebox{.45\linewidth}{!}{
\begin{tabular}{c|cccccccc}
\toprule
\multicolumn{1}{c}{\textbf{Finetuning }} &
\multicolumn{1}{c}{\textbf{MMLU}} & \multicolumn{1}{c}{\textbf{BBH}} & \textbf{GSM8K} & \multirow{2}{*}{\bf Avg.} \\
\textbf{Strategy} & Direct & Direct & CoT &  \\
\midrule
Baseline$_{\textsc{Flan-EC}_\textsc{base}}$  & 40.0 & 33.2 & 6.6 & 37.7  \\
\midrule
Freeze-Gate$_{\textsc{Flan-EC}_\textsc{base}}$  & 40.2 & 33.9 & 6.6 & 38.0 \\
Freeze-Expert$_{\textsc{Flan-EC}_\textsc{base}}$ & 38.3 & 32.5 & 5.4 & 36.2 \\
Freeze-MoE$_{\textsc{Flan-EC}_\textsc{base}}$  & 38.4 & 32.2 & 5.3 & 36.2 \\
\midrule
Z-loss$_{\textsc{Flan-EC}_\textsc{base}}$  & 38.9 & 32.8 & 5.7 & 36.8 \\
Balance-loss$_{\textsc{Flan-EC}_\textsc{base}}$  & 40.8 & 33.4 & 7.1 & 38.3 \\
\bottomrule
\end{tabular}} & 
\resizebox{.45\linewidth}{!}{
\begin{tabular}{c|cccccccc}
\toprule
\multicolumn{1}{c}{\textbf{Finetuning }} &
\multicolumn{1}{c}{\textbf{MMLU}} & \multicolumn{1}{c}{\textbf{BBH}} & \textbf{GSM8K} & \multirow{2}{*}{\bf Avg.} \\
\textbf{Strategy} & Direct & Direct & CoT &  \\
\midrule
Baseline$_{\textsc{Flan-ST}_\textsc{base}}$  & 40.1 & 33.3 & 6.4 & 37.8 \\
\midrule
Freeze-Gate$_{\textsc{Flan-ST}_\textsc{base}}$  & 40.6 & 33.5 & 6.4 & 38.2 \\
Freeze-Expert$_{\textsc{Flan-ST}_\textsc{base}}$ & 39.6 & 32.9 & 4.5 & 37.3 \\
Freeze-MoE$_{\textsc{Flan-ST}_\textsc{base}}$  & 39.2 & 32.9 & 3.6 & 36.9\\
\midrule
Z-loss$_{\textsc{Flan-ST}_\textsc{base}}$  & 40.6 & 33.4 & 6.5 & 38.1 \\
Balance-loss$_{\textsc{Flan-ST}_\textsc{base}}$  & 38.8 & 31.3 & 3.6 & 36.2 \\
\bottomrule
\end{tabular}
}
\vspace{1mm}
\end{tabular}
\caption{Ablations on different finetuning strategies of \textsc{Flan}-EC$_\textsc{base}$ and \textsc{Flan}-ST$_\textsc{base}$. 
}
\label{tbl:ablation_freeze}
\end{table}

\paragraph{Hyperparameter Sensitivity.} Following ST-MoE~\cite{stmoe}, we further experiment with expert dropout ($0.0, 0.1, 0.5$), varying the learning rate ($1e^{-4}, 5e^{-4}, 1e^{-3}$) and batch size ($16, 32, 64$) to examine the hyperparameter sensitivity of \shortname{}. We found that the performance varies in different tasks but not significantly with all the hyperparameters, but lower learning rate and small batch size lead to a more stable instruction finetuning process of the model at extra-large scales. 

\begin{figure*}[tb]
\centering
    \renewcommand\tabcolsep{1pt}
\begin{tabular}{cc}
\begin{subfigure}[b]{0.49\textwidth}
\includegraphics[width=\textwidth]{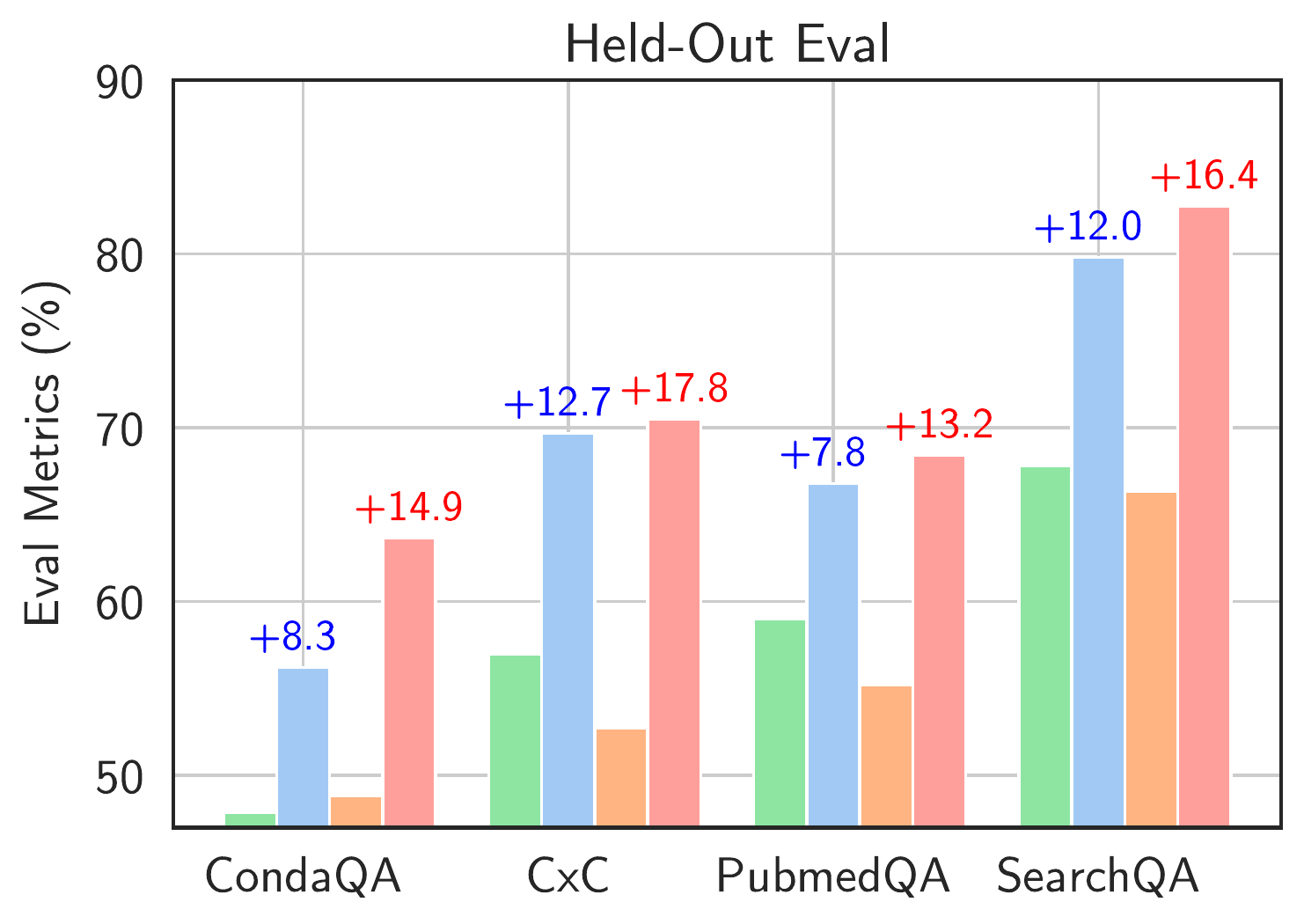}
\caption{$\textsc{Flan-EC}_\textsc{base}$ v.s. $\textsc{Flan-T5}_\textsc{base}$}
\end{subfigure} 
&
\begin{subfigure}[b]{0.49\textwidth}
\includegraphics[width=\textwidth]{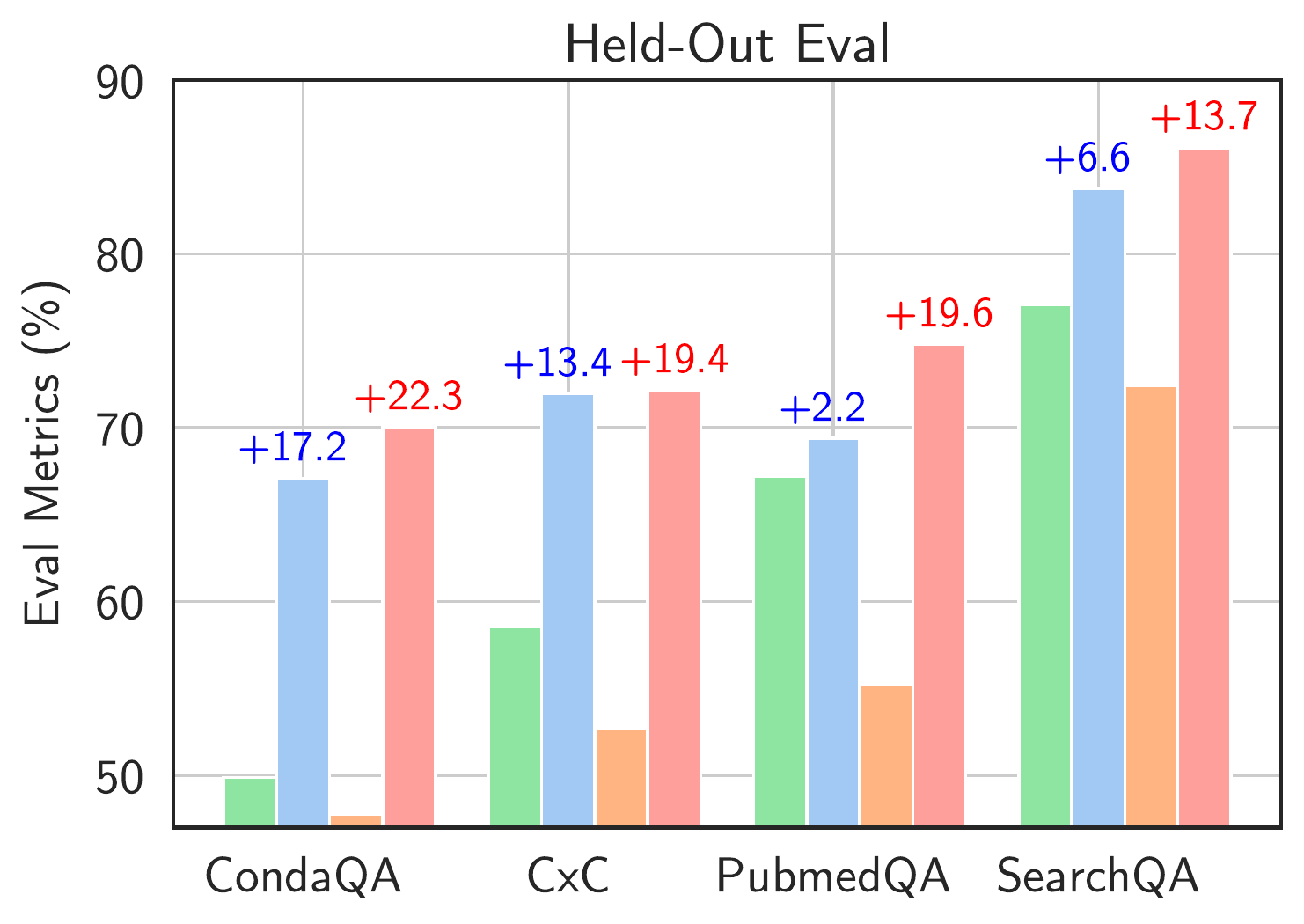}
\caption{$\textsc{Flan-EC}_\textsc{large}$ v.s. $\textsc{Flan-T5}_\textsc{large}$}
\end{subfigure}
\\
& \hspace{-70mm} \includegraphics[width=0.8\textwidth]{figs/finetune_gap_legend.pdf} 
\end{tabular}
\caption{\shortname{} Outperforms MoE on Single-Task Finetuning. We compare single-task finetuned MoE, single-task finetuned \shortname{}, and dense counterparts. The performance gap between \shortname{} and MoE is noticeably larger than that between FLAN-T5 and T5.}
\label{fig:finetune-gap-flan}
\end{figure*}

\paragraph{Finetuning v.s. Instruction Finetuning.} 
To compare the gap between finetuning MoE directly and \shortname{}, we experiment with single-task finetuned MoE, single-task finetuned \shortname{}, and dense counterparts in Figure~\ref{fig:finetune-gap-flan}. 
We perform hyper-parameter search for each finetuning setting. 

For the examined Held-Out tasks, we observed that the improvement of \shortname{} over finetuning MoE is noticeably larger compared to the performance gap between \sshortname{}-T5 and T5. 
This difference becomes even more pronounced when there is a scarcity of labeled data or when the model size is increased. These observations confirm the benefits of \shortname{} in mitigating overfitting issues associated with directly finetuning MoE. 

Despite their advantages such as increased adaptability and efficiency in managing complex tasks, MoE architectures are prone to overfitting during the finetuning process, as discussed in citation. This can be seen in Figures~\ref{fig:finetune-gap-flan} and ~\ref{fig:main-finetune-gap-flan}, where single-task fine-tuned MoE models sometimes underperform their dense T5 counterparts.

Interestingly, compared to dense models, MoE models derive greater benefits from instruction-tuning and are more sensitive to the number of instruction-tuning tasks. In general, MoE model performance scales better with respect to the number of tasks rather than the number of experts. We hypothesize this is primarily due to the specialized nature of individual experts, which can lead to heightened sensitivity to noise and limited generalization capabilities when exposed to unseen data.


\subsection{Additional Analysis}

\begin{wrapfigure}[15]{r}{0.45\textwidth}
    \vspace{-1.05cm}
    \includegraphics[width=0.45\textwidth]{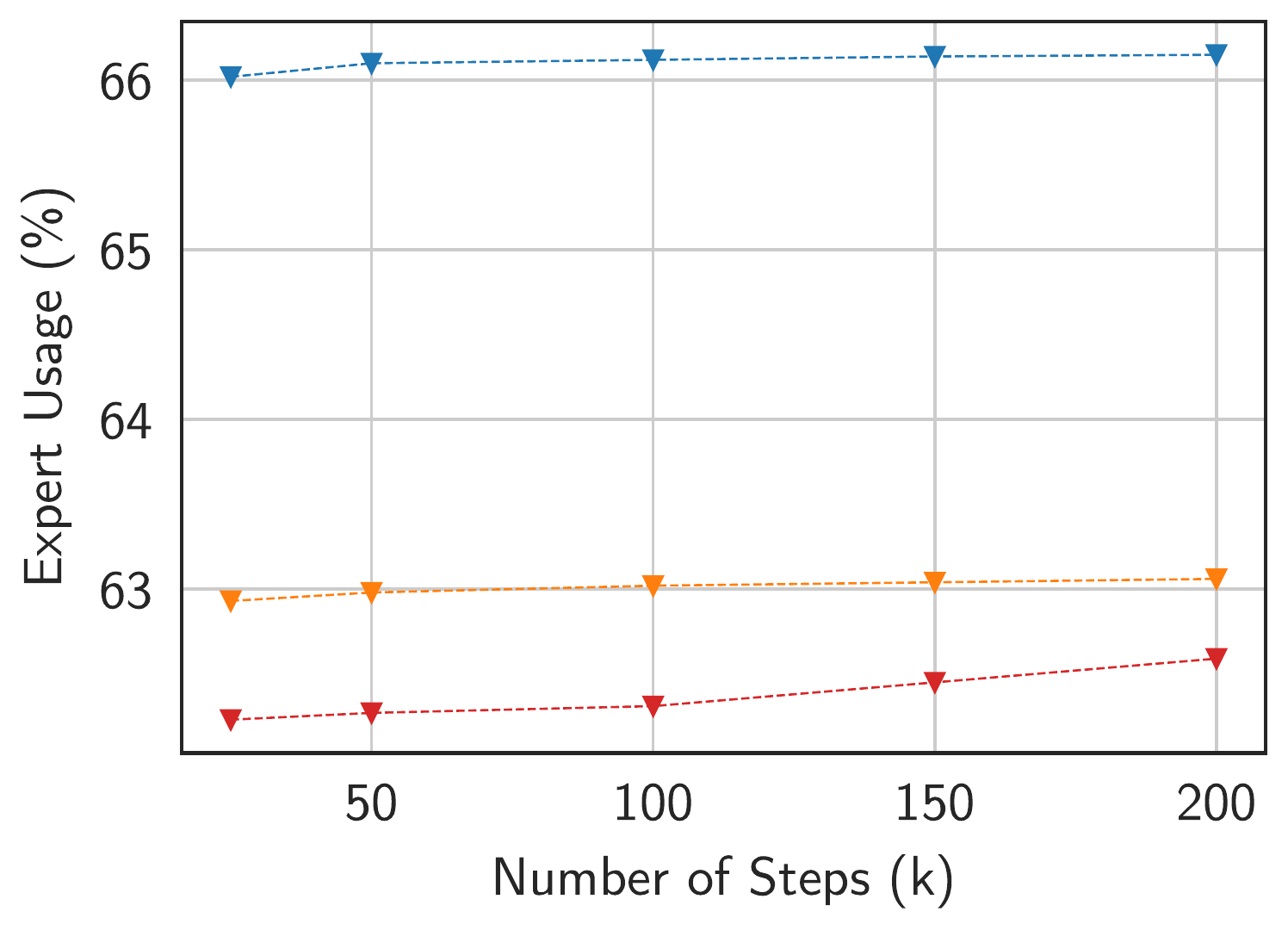}
    \caption{ Expert usage of \textsc{Flan-EC} at different scales during instruction finetuning, where larger models entail smaller expert usage. 
    }
    \label{fig:expert_usage}
\end{wrapfigure}

\paragraph{Expert Specialization.} 
As the size of a \shortname{} model increases in Figure~\ref{fig:expert_usage}, a notable rise in expert specialization tends to occur. 
Larger models entail a higher number of parameters and more complex structures, which inherently provide a broader scope for each expert to specialize in specific facets of the problem space. This increased specialization can be understood as a form of division of labor, where each expert sub-network becomes adept at handling a certain type of task or data pattern. Consequently, the overall model can demonstrate a higher degree of adaptability and precision in tackling diverse and complex tasks. 
We also observe that after instruction-tuning, the MoE models exhibit better expert usage, which may help prevent the expert collapse for generalization after instruction-tuning as in~\cite{zuo2021taming}. 

\paragraph{Failure Cases.} 
The fine-grained specialization of \shortname{} models, particularly when fine-tuned on English-only instructions, can inadvertently lead to a narrowing of the model's capacity to effectively process and generate content in multiple languages. 
We found all the \shortname{} perform poorly on multi-lingual benchmarks including TyDiQA and MGSM. 
Even the largest \sshortname{}-ST$_\textsc{32b}$ only achieves 15.5\% on MGSM and 25.1\% on TyDiQA, which is only comparable to the vanilla \palm$_\textsc{62B}$ with 18.2\% on MSGM, and \palm$_\textsc{8B}$ with 25.0\% on TyDiQA. It also underperform \flanpalm variants. 
We hypotheses that 
this issue may stes from the model's over-optimization towards the specificities of the English language during finetuning, which can impede its ability to navigate the complexities of other languages. Consequently, while MoE models offer significant benefits in terms of task-specific adaptability and efficiency, their potential shortcomings in multilinguality highlight the importance of incorporating diverse linguistic data during the training process to ensure broad and effective language coverage. 

\section{Related Work}
\label{sec:related_work}

\paragraph{Instruction Tuning.} 
Instruction tuning has evolved as a strategy to enhance the functionality and interactivity of large language models (LLMs) for dialogues and complex tasks. 
Prior studies, including~\cite{t5,mtdnn,ext5}, have delved into large-scale multi-task fine-tuning to enhance the downstream single target fine-tuning, albeit without instruction prompts. 
Initiatives such as UnifiedQA~\cite{unifiedqa,mccann2018natural,keskar2019unifying} have amalgamated a multitude of NLP tasks into a singular generative question answering format, utilizing prompt instructions for multi-task fine-tuning and evaluation. 

Efforts like Natural Instructions~\cite{mishra2021cross}, Flan 2021~\cite{flan}, and P3 (the Public Pool of Prompts, ~\cite{t0}) have collated vast NLP task collections, templatizing them with instructions for fine-tuning models to enhance their adaptability to unseen instructions. Some studies, such as Super-Natural Instructions~\cite{naturalinstructions} and OPT-IML~\cite{iyer2022opt}, took this a step further by combining numerous datasets and tasks into a single resource. In the meantime, others like xP3~\cite{muennighoff2022crosslingual} introduced multilingual instruction tuning and Flan 2022 [4] employed Chain-of-Thought training prompts.

Recently, there has been a move towards expanding task diversity more assertively using synthetic data generation, particularly for creative and open-ended dialogue ~\cite{wang2022self,honovich2022unnatural,ye2021crossfit}. 
Some researchers have also tried to provide human feedback on language model responses ~\cite{ouyang2022training,glaese2022improving,nakano2021webgpt,bai2022constitutional,bai2022training}, or bridge the modality gap with multi-modal instruction fine-tuning~\cite{liu2023visual,dai2023instructblip,li2023otter}.

\paragraph{Sparse Mixture of Experts models.} 
The foundation of our work is built on the concept of deep sparse Mixture-of-Experts (MoEs), a topic that has been independently explored in both Computer Vision~\cite{vmoe,mixermoe,limoe,vlmoe} and Natural Language Processing~\cite{mixermoe,limoe,moe,gshard,switchtransformer,glam,stmoe,unifiedscaling,expertchoice,sparseupcycle,taskmoe,zuo2021taming}. The idea revolves around conditional computation, which aims to enhance the number of model parameters without a corresponding rise in computational expense. This is achieved by selectively activating only the relevant portions of the model, based on input-dependent factors. 
MoE models leverage a learned gating mechanism that triggers only a select subset of $k$ experts out of a total of $E$ for a given input. This approach allows an input to either select all experts~\cite{eigen2013learning} or merely a sparse mixture of them, as observed in recent massive language models~\cite{switchtransformer,glam}.
While a number of studies have sought to enhance the gating mechanism itself~\cite{hazimeh2021dselectk,lewis2021base,roller2021hash,expertchoice}, MoE models have also been explored in the context of multitask learning~\cite{hazimeh2021dselectk,taskmoe}. Typically, a shared pool of experts is used, although there has been investigation into per-task routers~\cite{mmoe}. This essentially permits an input to choose the most relevant expert(s) for a given task, thereby optimizing the processing and results. 
Nevertheless, the instability of MoE models during fine-tuning or multitask learning has consistently been a challenge. Our study aims to investigate whether instruction fine-tuning with scaled tasks might contribute to mitigating the generalization issues inherent to MoE models. 
\section{Conclusion}
\label{sec:conclusion}

In this work, we have introduced \shortname{}, an innovative method to amplify the scalability of instruction-tuned language models by employing the sparse Mixture-of-Experts (MoE) technique. Our strategy amalgamates the merits of instruction-finetuning, which bolsters task-specific performance, and MoE, which provides computational efficiency coupled with diminished memory requirements.

We have substantiated the effectiveness of \shortname{} through comprehensive experiments across a wide spectrum of Natural Language Processing (NLP) tasks, such as natural language understanding, question answering, and reasoning. Our results consistently underscore the superior performance of \shortname{} over current state-of-the-art methods, marking substantial advancements in both accuracy and efficiency. Notably, these advancements are attained without necessitating an increase in computational resources or memory usage during training and inference, often even reducing the resource requirements in the process.

{\small
\bibliography{egbib}
\bibliographystyle{plain}
}

 \newpage

\appendix
\noindent\makebox[\linewidth]{\rule{\linewidth}{3.5pt}}
\begin{center}
\bf{\Large Appendix for \\
``Mixture-of-Experts Meets Instruction Tuning: A Winning Combination for Large Language Models''}
\end{center}
\noindent\makebox[\linewidth]{\rule{\linewidth}{1pt}}

\section{Full Experiment Results}
\subsection{MMLU}
In the case of five-shot MMLU, we employ the "dev" set as the small sample exemplars. The performance of individual tasks in MMLU on the "validation" set is detailed in this section (refer to \url{https://www.tensorflow.org/datasets/community_catalog/huggingface/hendrycks_test} for more information). Please note, all MMLU findings presented in this paper correspond to the "validation" set. 
We employ the prompts in ~\cite{flant5}.

\begin{table}[h]
\centering
\caption{MMLU[:10] individual task performance.}
\label{tab:my-table}
\setlength{\tabcolsep}{3pt}
\resizebox{\columnwidth}{!}{%
\begin{tabular}{llcccccccccccccccccccc}
\toprule
& \multicolumn{11}{c}\thead{MMLU} \\
\cmidrule(lr){3-22}
 &
  & \multicolumn{2}{c}{\thead{Abstract \\Algebra}} &
  \multicolumn{2}{c}{\thead{Anatomy}} &
  \multicolumn{2}{c}{\thead{Astronomy}} &
  \multicolumn{2}{c}{\thead{Business\\Ethics}} &
  \multicolumn{2}{c}{\thead{Clinical\\Knowledge}} &
  \multicolumn{2}{c}{\thead{College\\Biology}} &
  \multicolumn{2}{c}{\thead{College\\Chemistry}} &
  \multicolumn{2}{c}{\thead{College\\Comp. Sci.}} &
  \multicolumn{2}{c}{\thead{College\\Math}} &
  \multicolumn{2}{c}{\thead{College\\Medicine}} \\
  \cmidrule(lr){3-4}
  \cmidrule(lr){5-6}
  \cmidrule(lr){7-8}
  \cmidrule(lr){9-10}
  \cmidrule(lr){11-12}
  \cmidrule(lr){13-14}
  \cmidrule(lr){15-16}
  \cmidrule(lr){17-18}
  \cmidrule(lr){19-20}
  \cmidrule(lr){21-22} 
 \multicolumn{2}{l}{Model} 
  & \thead{Direct} &
  \thead{CoT} &
  \thead{Direct} &
  \thead{CoT} &
  \thead{Direct} &
  \thead{CoT} &
  \thead{Direct} &
  \thead{CoT} &
  \thead{Direct} &
  \thead{CoT} &
  \thead{Direct} &
  \thead{CoT} &
  \thead{Direct} &
  \thead{CoT} &
  \thead{Direct} &
  \thead{CoT} &
  \thead{Direct} &
  \thead{CoT} &
  \thead{Direct} &
  \thead{CoT} \\
  \midrule
- & davinci  & 27.3 & 27.3 & 50.0 & 42.9 & 25.0 & 31.2 & 45.5 & 36.4 & 31.0 & 34.5 & 43.8 & 25.0 & 12.5 & 25.0 & 18.2 & 36.4 & 27.3 & 9.1 & 36.4 & 31.8 \\
- & text-davinci-002  & 9.1 & 27.3 & 57.1 & 28.6 & 62.5 & 56.2 & 63.6 & 72.7 & 51.7 & 55.2 & 68.8 & 43.8 & 12.5 & 37.5 & 63.6 & 36.4 & 54.5 & 36.4 & 63.6 & 54.5 \\
- & text-davinci-003 & 18.2 & 36.4 & 50.0 & 57.1 & 62.5 & 62.5 & 63.6 & 63.6 & 62.1 & 65.5 & 62.5 & 81.2 & 25.0 & 25.0 & 54.5 & 45.5 & 81.8 & 72.7 & 72.7 & 68.2 \\ 
- & code-davinci-002  & 18.2 & 27.3 & 71.4 & 35.7 & 68.8 & 56.2 & 54.5 & 63.6 & 69.0 & 65.5 & 62.5 & 50.0 & 25.0 & 37.5 & 45.5 & 27.3 & 72.7 & 45.5 & 77.3 & 86.4 \\
\bottomrule
\end{tabular}%
}
\end{table}

\begin{table}[]
\centering
\caption{MMLU[10:20] individual task performance.}
\label{tab:my-table}
\setlength{\tabcolsep}{3pt}
\resizebox{\columnwidth}{!}{%
\begin{tabular}{llcccccccccccccccccccc}
\toprule
& \multicolumn{11}{c}\thead{MMLU} \\
\cmidrule(lr){3-22}
 &
  & \multicolumn{2}{c}{\thead{College\\Physics}} & \multicolumn{2}{c}{\thead{Computer\\Security}} & \multicolumn{2}{c}{\thead{Conceptual\\physics}} & \multicolumn{2}{c}{\thead{Econometrics}} & \multicolumn{2}{c}{\thead{Electrical\\Engineering}} & \multicolumn{2}{c}{\thead{Elementary\\Mathematics}} & \multicolumn{2}{c}{\thead{Formal\\Logic}} & \multicolumn{2}{c}{\thead{Global\\Facts}} & \multicolumn{2}{c}{\thead{High School\\Biology}} & \multicolumn{2}{c}{\thead{High School\\Chemistry}} \\
  \cmidrule(lr){3-4}
  \cmidrule(lr){5-6}
  \cmidrule(lr){7-8}
  \cmidrule(lr){9-10}
  \cmidrule(lr){11-12}
  \cmidrule(lr){13-14}
  \cmidrule(lr){15-16}
  \cmidrule(lr){17-18}
  \cmidrule(lr){19-20}
  \cmidrule(lr){21-22} 
 \multicolumn{2}{l}{Model} 
  & \thead{Direct} &
  \thead{CoT} &
  \thead{Direct} &
  \thead{CoT} &
  \thead{Direct} &
  \thead{CoT} &
  \thead{Direct} &
  \thead{CoT} &
  \thead{Direct} &
  \thead{CoT} &
  \thead{Direct} &
  \thead{CoT} &
  \thead{Direct} &
  \thead{CoT} &
  \thead{Direct} &
  \thead{CoT} &
  \thead{Direct} &
  \thead{CoT} &
  \thead{Direct} &
  \thead{CoT} \\
  \midrule
- & davinci & 45.5 & 36.4 & 72.7 & 54.5 & 38.5 & 46.2 & 25.0 & 33.3 & 25.0 & 50.0 & 24.4 & 29.3 & 14.3 & 14.3 & 20.0 & 20.0 & 28.1 & 34.4 & 31.8 & 13.6 \\
- & text-davinci-002  & 54.5 & 81.8 & 81.8 & 81.8 & 53.8 & 61.5 & 58.3 & 50.0 & 50.0 & 37.5 & 56.1 & 73.2 & 7.1 & 28.6 & 50.0 & 70.0 & 71.9 & 71.9 & 18.2 & 36.4 \\
 - & text-davinci-003 & 36.4 & 45.5 & 81.8 & 63.6 & 42.3 & 57.7 & 58.3 & 58.3 & 50.0 & 56.2 & 48.8 & 75.6 & 42.9 & 42.9 & 40.0 & 50.0 & 71.9 & 75.0 & 36.4 & 36.4 \\ \vspace{3mm} 
- & code-davinci-002 & 45.5 & 72.7 & 90.9 & 81.8 & 53.8 & 57.7 & 66.7 & 41.7 & 50.0 & 50.0 & 56.1 & 75.6 & 50.0 & 42.9 & 40.0 & 50.0 & 71.9 & 65.6 & 40.9 & 40.9 \\
\bottomrule
\end{tabular}%
}
\end{table}

\begin{table}[]
\centering
\caption{MMLU[20:30] individual task performance.}
\label{tab:my-table}
\setlength{\tabcolsep}{3pt}
\resizebox{\columnwidth}{!}{%
\begin{tabular}{llcccccccccccccccccccc}
\toprule
& \multicolumn{11}{c}\thead{MMLU} \\
\cmidrule(lr){3-22}
 &
  & \multicolumn{2}{c}{\thead{High School\\Comp. Sci.}} & \multicolumn{2}{c}{\thead{High School\\European History}} & \multicolumn{2}{c}{\thead{High School\\Geography}} & \multicolumn{2}{c}{\thead{High School\\Gvmt \& Politics}} & \multicolumn{2}{c}{\thead{High School\\Macroeconomics}} & \multicolumn{2}{c}{\thead{High School\\Math}} & \multicolumn{2}{c}{\thead{High School\\Microeconomics}} & \multicolumn{2}{c}{\thead{High School\\Physics}} & \multicolumn{2}{c}{\thead{High School\\Psychology}} & \multicolumn{2}{c}{\thead{High School\\Statistics}} \\
  \cmidrule(lr){3-4}
  \cmidrule(lr){5-6}
  \cmidrule(lr){7-8}
  \cmidrule(lr){9-10}
  \cmidrule(lr){11-12}
  \cmidrule(lr){13-14}
  \cmidrule(lr){15-16}
  \cmidrule(lr){17-18}
  \cmidrule(lr){19-20}
  \cmidrule(lr){21-22} 
 \multicolumn{2}{l}{Model} 
  & \thead{Direct} &
  \thead{CoT} &
  \thead{Direct} &
  \thead{CoT} &
  \thead{Direct} &
  \thead{CoT} &
  \thead{Direct} &
  \thead{CoT} &
  \thead{Direct} &
  \thead{CoT} &
  \thead{Direct} &
  \thead{CoT} &
  \thead{Direct} &
  \thead{CoT} &
  \thead{Direct} &
  \thead{CoT} &
  \thead{Direct} &
  \thead{CoT} &
  \thead{Direct} &
  \thead{CoT} \\
  \midrule
- & davinci & 55.6 & 44.4 & 38.9 & 33.3 & 63.6 & 63.6 & 52.4 & 52.4 & 39.5 & 51.2 & 13.8 & 10.3 & 34.6 & 46.2 & 29.4 & 11.8 & 50.0 & 65.0 & 34.8 & 26.1 \\
- & text-davinci-002 & 100.0 & 66.7 & 83.3 & 83.3 & 81.8 & 77.3 & 76.2 & 76.2 & 62.8 & 74.4 & 34.5 & 24.1 & 76.9 & 73.1 & 47.1 & 23.5 & 88.3 & 90.0 & 52.2 & 43.5 \\
- & text-davinci-003 & 66.7 & 55.6 & 83.3 & 77.8 & 95.5 & 77.3 & 81.0 & 81.0 & 67.4 & 62.8 & 44.8 & 51.7 & 80.8 & 76.9 & 29.4 & 23.5 & 95.0 & 91.7 & 52.2 & 52.2 \\ \vspace{3mm}
- & code-davinci-002  & 88.9 & 55.6 & 83.3 & 77.8 & 90.9 & 86.4 & 85.7 & 85.7 & 67.4 & 67.4 & 48.3 & 51.7 & 88.5 & 80.8 & 23.5 & 29.4 & 95.0 & 90.0 & 65.2 & 65.2 \\
\bottomrule
\end{tabular}%
}
\end{table}

\begin{table}[]
\centering
\caption{MMLU[30:40] individual task performance.}
\label{tab:my-table}
\setlength{\tabcolsep}{3pt}
\resizebox{\columnwidth}{!}{%
\begin{tabular}{llcccccccccccccccccccc}
\toprule
& \multicolumn{11}{c}\thead{MMLU} \\
\cmidrule(lr){3-22}
 &
 &\multicolumn{2}{c}{\thead{High School\\US History}} &
  \multicolumn{2}{c}{\thead{High School\\World History}} &
  \multicolumn{2}{c}{\thead{Human\\Aging}} &
  \multicolumn{2}{c}{\thead{Human\\Sexuality}} &
  \multicolumn{2}{c}{\thead{International\\Law}} &
  \multicolumn{2}{c}{\thead{Jurisprudence}} &
  \multicolumn{2}{c}{\thead{Logical\\Fallacies}} &
  \multicolumn{2}{c}{\thead{Machine\\Learning}} &
  \multicolumn{2}{c}{\thead{Management}} &
  \multicolumn{2}{c}{\thead{Marketing}} \\
  \cmidrule(lr){3-4}
  \cmidrule(lr){5-6}
  \cmidrule(lr){7-8}
  \cmidrule(lr){9-10}
  \cmidrule(lr){11-12}
  \cmidrule(lr){13-14}
  \cmidrule(lr){15-16}
  \cmidrule(lr){17-18}
  \cmidrule(lr){19-20}
  \cmidrule(lr){21-22} 
 \multicolumn{2}{l}{Model} 
  & \thead{Direct} &
  \thead{CoT} &
  \thead{Direct} &
  \thead{CoT} &
  \thead{Direct} &
  \thead{CoT} &
  \thead{Direct} &
  \thead{CoT} &
  \thead{Direct} &
  \thead{CoT} &
  \thead{Direct} &
  \thead{CoT} &
  \thead{Direct} &
  \thead{CoT} &
  \thead{Direct} &
  \thead{CoT} &
  \thead{Direct} &
  \thead{CoT} &
  \thead{Direct} &
  \thead{CoT} \\
  \midrule
- & davinci & 54.5 & 36.4 & 38.5 & 46.2 & 30.4 & 60.9 & 16.7 & 50.0 & 84.6 & 38.5 & 18.2 & 9.1 & 55.6 & 50.0 & 27.3 & 18.2 & 45.5 & 63.6 & 56.0 & 64.0 \\
- & text-davinci-002  & 86.4 & 72.7 & 69.2 & 73.1 & 78.3 & 87.0 & 66.7 & 58.3 & 92.3 & 84.6 & 63.6 & 45.5 & 77.8 & 66.7 & 45.5 & 36.4 & 72.7 & 72.7 & 80.0 & 80.0 \\
- & text-davinci-003 & 81.8 & 81.8 & 80.8 & 76.9 & 78.3 & 73.9 & 66.7 & 58.3 & 84.6 & 84.6 & 63.6 & 54.5 & 83.3 & 83.3 & 45.5 & 54.5 & 81.8 & 72.7 & 84.0 & 76.0 \\ \vspace{3mm}
- & code-davinci-002  & 100.0 & 77.3 & 76.9 & 84.6 & 78.3 & 78.3 & 75.0 & 58.3 & 100.0 & 92.3 & 63.6 & 72.7 & 83.3 & 72.2 & 54.5 & 63.6 & 90.9 & 81.8 & 80.0 & 80.0  \\
\bottomrule
\end{tabular}%
}
\end{table}

\begin{table}[]
\centering
\caption{MMLU[40:50] individual task performance.}
\label{tab:my-table}
\setlength{\tabcolsep}{3pt}
\resizebox{\columnwidth}{!}{%
\begin{tabular}{llcccccccccccccccccccc}
\toprule
& \multicolumn{11}{c}\thead{MMLU} \\
\cmidrule(lr){3-22}
 &
 &\multicolumn{2}{c}{\thead{Medical\\Genetics}} &
  \multicolumn{2}{c}{\thead{Misc.}} &
  \multicolumn{2}{c}{\thead{Moral\\Disputes}} &
  \multicolumn{2}{c}{\thead{Moral\\Scenarios}} &
  \multicolumn{2}{c}{\thead{Nutrition}} &
  \multicolumn{2}{c}{\thead{Philosophy}} &
  \multicolumn{2}{c}{\thead{Prehistory}} &
  \multicolumn{2}{c}{\thead{Professional\\Accounting}} &
  \multicolumn{2}{c}{\thead{Professional\\Law}} &
  \multicolumn{2}{c}{\thead{Professional\\Medicine}} \\
  \cmidrule(lr){3-4}
  \cmidrule(lr){5-6}
  \cmidrule(lr){7-8}
  \cmidrule(lr){9-10}
  \cmidrule(lr){11-12}
  \cmidrule(lr){13-14}
  \cmidrule(lr){15-16}
  \cmidrule(lr){17-18}
  \cmidrule(lr){19-20}
  \cmidrule(lr){21-22} 
 \multicolumn{2}{l}{Model} 
  & \thead{Direct} &
  \thead{CoT} &
  \thead{Direct} &
  \thead{CoT} &
  \thead{Direct} &
  \thead{CoT} &
  \thead{Direct} &
  \thead{CoT} &
  \thead{Direct} &
  \thead{CoT} &
  \thead{Direct} &
  \thead{CoT} &
  \thead{Direct} &
  \thead{CoT} &
  \thead{Direct} &
  \thead{CoT} &
  \thead{Direct} &
  \thead{CoT} &
  \thead{Direct} &
  \thead{CoT} \\
  \midrule
- & davinci  & 72.7 & 90.9 & 50.0 & 65.1 & 57.9 & 39.5 & 24.0 & 34.0 & 54.5 & 45.5 & 44.1 & 61.8 & 45.7 & 42.9 & 29.0 & 35.5 & 31.2 & 26.5 & 32.3 & 38.7 \\
- & text-davinci-002  & 90.9 & 90.9 & 79.1 & 81.4 & 63.2 & 65.8 & 46.0 & 40.0 & 75.8 & 69.7 & 67.6 & 67.6 & 60.0 & 65.7 & 64.5 & 41.9 & 45.3 & 38.8 & 64.5 & 71.0 \\
- & text-davinci-003 & 100.0 & 100.0 & 82.6 & 87.2 & 71.1 & 52.6 & 43.0 & 65.0 & 78.8 & 69.7 & 76.5 & 76.5 & 65.7 & 74.3 & 54.8 & 38.7 & 48.8 & 47.1 & 74.2 & 67.7 \\ \vspace{3mm}
- & code-davinci-002  & 100.0 & 100.0 & 84.9 & 87.2 & 68.4 & 50.0 & 41.0 & 60.0 & 69.7 & 66.7 & 79.4 & 76.5 & 77.1 & 77.1 & 51.6 & 51.6 & 54.7 & 38.2 & 77.4 & 80.6 \\
\bottomrule
\end{tabular}%
}
\end{table}

\begin{table}[]
\centering
\caption{MMLU[50:57] individual task performance.}
\label{tab:my-table}
\setlength{\tabcolsep}{3pt}
\resizebox{\columnwidth}{!}{%
\begin{tabular}{llcccccccccccccccc}
\toprule
& \multicolumn{10}{c}\thead{MMLU} \\
\cmidrule(lr){3-18}
 &
  &\multicolumn{2}{c}{\thead{Professional\\Psychology}} &
  \multicolumn{2}{c}{\thead{Public\\Relations}} &
  \multicolumn{2}{c}{\thead{Security\\Studies}} &
  \multicolumn{2}{c}{\thead{Sociology}} &
  \multicolumn{2}{c}{\thead{US Foreign\\Policy}} &
  \multicolumn{2}{c}{\thead{Virology}} &
  \multicolumn{2}{c}{\thead{World Religions}} &
  \multicolumn{2}{c}{\thead{\textbf{Average}}} \\
  \cmidrule(lr){3-4}
  \cmidrule(lr){5-6}
  \cmidrule(lr){7-8}
  \cmidrule(lr){9-10}
  \cmidrule(lr){11-12}
  \cmidrule(lr){13-14}
  \cmidrule(lr){15-16}
  \cmidrule(lr){17-18}
 \multicolumn{2}{l}{Model} 
  & \thead{Direct} &
  \thead{CoT} &
  \thead{Direct} &
  \thead{CoT} &
  \thead{Direct} &
  \thead{CoT} &
  \thead{Direct} &
  \thead{CoT} &
  \thead{Direct} &
  \thead{CoT} &
  \thead{Direct} &
  \thead{CoT} &
  \thead{Direct} &
  \thead{CoT} &
  \thead{Direct} &
  \thead{CoT} \\
  \midrule
- & davinci  & 37.7 & 43.5 & 50.0 & 50.0 & 44.4 & 40.7 & 63.6 & 59.1 & 45.5 & 63.6 & 33.3 & 27.8 & 63.2 & 68.4 & 39.7 & 40.5 \\
- & text-davinci-002 & 65.2 & 58.0 & 50.0 & 50.0 & 77.8 & 48.1 & 90.9 & 86.4 & 81.8 & 81.8 & 44.4 & 33.3 & 84.2 & 78.9 & 63.1 & 60.0 \\
- & text-davinci-003 & 68.1 & 63.8 & 50.0 & 50.0 & 70.4 & 63.0 & 86.4 & 95.5 & 81.8 & 90.9 & 50.0 & 50.0 & 84.2 & 84.2 & 64.8	& 64.6 \\ \vspace{3mm} 
- & code-davinci-002  & 76.8 & 66.7 & 50.0 & 58.3 & 74.1 & 51.9 & 86.4 & 90.9 & 90.9 & 72.7 & 50.0 & 44.4 & 84.2 & 78.9 & 68.2 & 64.5 \\
80M & T5-Small &  20.3   &   4.3   & 33.3   &  16.7   & 18.5   &   0.0   & 22.7   &   0.0   & 27.3   &   9.1   & 27.8   &   5.6   & 21.1   &  15.8  & 26.7 & 5.6  \\
\bottomrule
\end{tabular}%
}
\end{table}

\clearpage
\newpage
\subsection{BBSH}
BBH refers to a subset of difficult tasks from BIG-Bench, handpicked by ~\cite{suzgun2022challenging} in 2022, where the model proposed by~\cite{bigbench} in the same year outperformed the average human rater. ~\cite{suzgun2022challenging} mentions 23 tasks, two of which consist of three subtasks each. For ease of interpretation, we treat these subtasks as standalone tasks and calculate an unweighted average. We utilize the prompts provided in ~\cite{suzgun2022challenging}'s study. 

\begin{table}[h]
\centering
\caption{BBH[:9] individual task performance.}
\label{tab:my-table}
\setlength{\tabcolsep}{3pt}
\resizebox{\columnwidth}{!}{%
\begin{tabular}{llcccccccccccccccccc}
\toprule
& & \multicolumn{18}{c}{BBH}  \\
\cmidrule(lr){3-20}
 &
  & \multicolumn{2}{c}{\thead{Boolean\\Expressions}} &
  \multicolumn{2}{c}{\thead{Causal\\Judgement}} &
  \multicolumn{2}{c}{\thead{Date\\Understanding}} &
  \multicolumn{2}{c}{\thead{Disambiguation\\QA}} &
  \multicolumn{2}{c}{\thead{Dyck\\Languages}} &
  \multicolumn{2}{c}{\thead{Formal\\Fallacies}} &
  \multicolumn{2}{c}{\thead{Geometric\\Shapes}} &
  \multicolumn{2}{c}{\thead{Hyperbaton}} &
  \multicolumn{2}{c}{\thead{Logical Deduction\\Five Objects}} \\
  \cmidrule(lr){3-4}
  \cmidrule(lr){5-6}
  \cmidrule(lr){7-8}
  \cmidrule(lr){9-10}
  \cmidrule(lr){11-12}
  \cmidrule(lr){13-14}
  \cmidrule(lr){15-16}
  \cmidrule(lr){17-18}
  \cmidrule(lr){19-20}
 \multicolumn{2}{l}{Model} 
  & \thead{Direct} &
  \thead{CoT} &
  \thead{Direct} &
  \thead{CoT} &
  \thead{Direct} &
  \thead{CoT} &
  \thead{Direct} &
  \thead{CoT} &
  \thead{Direct} &
  \thead{CoT} &
  \thead{Direct} &
  \thead{CoT} &
  \thead{Direct} &
  \thead{CoT} &
  \thead{Direct} &
  \thead{CoT} &
  \thead{Direct} &
  \thead{CoT} \\
  \midrule
- & davinci & 54.0 & 69.2 & 57.8 & 48.1 & 37.6 & 52.4 & 40.0 & 40.8 & 28.0 & 0.0 & 47.2 & 52.8 & 10.4 & 10.8 & 49.6 & 47.6 & 24.4 & 34.4 \\
- & text-davinci-002 & 90.0 & 87.6 & 57.8 & 56.1 & 55.6 & 81.6 & 66.4 & 70.8 & 42.0 & 32.0 & 52.4 & 58.4 & 35.2 & 56.0 & 67.2 & 72.4 & 31.6 & 51.2 \\
- & text-davinci-003 & 90.0	& 90.8 & 63.6 & 63.6 & 58.8 & 82.0 & 68.4 & 66.8 & 14.8 & 40.0 & 58.0 & 55.2 & 36.8 & 60.4 & 60.8 & 53.2 & 44.0 & 58.0 \\ \vspace{3mm}
- & code-davinci-002 & 88.4 & 92.8 & 63.6 & 54.0 & 63.6 & 87.2 & 67.2 & 76.0 & 46.8 & 56.8 & 52.4 & 50.4 & 32.0 & 54.4 & 60.4 & 66.4 & 32.4 & 54.8 \\
80M & T5-Small &  40.0   &   0.0   & 51.3   &   2.7   & 20.0   &  10.8   & 34.8   &  14.0   &  2.4   &   0.0   & 52.8   &   0.0   &  8.4   &   0.0   & 52.0   &   0.0   & 17.2   &   7.6  \\\vspace{3mm} 
& Flan-T5-Small &  54.0   &  39.6   & 48.1   &  42.8   & 22.4   &  20.4   & 31.2   &   2.0   &  0.0   &   0.0   & 53.2   &  46.8   &  8.8   &   4.0   & 65.2   &  13.2   & 22.0   &  19.2   \\
250M & T5-Base & 46.0   &  45.6   & 51.9   &  38.0   & 20.0   &  19.6   & 33.6   &  30.8   &  1.6   &   0.0   & 46.8   &  31.2   & 22.0   &   0.0   & 51.2   &   0.0   & 19.2   &   9.6    \\ \vspace{3mm} 
& Flan-T5-Base & 48.4 &46.4 &52.4 &47.1 &18.0 &20.4 &54.8 &44.8 &7.6 &0.0 &53.2 &49.2 &0.4 &12.8 &67.6 &58.8 &27.2 &22.0 \\
780M & T5-Large & 46.0   &  49.2   & 51.9   &  26.2   & 20.8   &  20.0   & 34.8   &  10.8   &  0.4   &   0.0   & 46.8   &   6.0   & 29.6   &   0.0   & 50.0   &   0.0   & 19.6   &  14.8  \\\vspace{3mm} 
& Flan-T5-Large &  64.0 &58.0 &56.1 &20.9 &24.4 &26.8 &67.6 &61.2 &0.8 &0.0 &22.8 &39.6 &0.8 &8.0 &72.4 &56.0 &47.6 &22.4  \\
3B & T5-XL &  55.2   &  47.2   & 52.4   &  26.7   & 21.6   &  22.4   & 32.4   &   4.8   &  6.0   &   0.0   & 47.2   &   7.2   &  8.4   &   0.0   & 52.0   &   0.0   & 22.0   &  22.8 \\\vspace{3mm} 
& Flan-T5-XL &  52.4 &56.0 &62.0 &56.1 &46.8 &48.8 &70.0 &70.4 &0.0 &0.0 &56.4 &48.0 &15.2 &4.4 &55.6 &56.8 &54.0 &32.4   \\
11B & T5-XXL & 49.6   &  65.2   & 52.4   &   1.6   & 35.2   &  54.0   & 35.2   &   0.0   &  2.0   &   0.0   & 52.4   &   0.0   & 15.6   &   0.0   & 55.6   &   0.0   & 18.0   &  37.2    \\\vspace{3mm} 
& Flan-T5-XXL  &56.8 &60.8 &60.4 &53.5 &69.6 &53.6 &71.2 &71.2 &0.8 &0.4 &55.6 &46.4 &14.0 &24.8 &71.6 &53.2 &55.6 &46.4  \\
8B & Flan-PaLM &  48.8   &  52.8   & 60.4   &  54.0   & 10.8   &  28.8   & 58.0   &  55.6   & 20.8   &   0.0   & 52.0   &  50.8   & 15.6   &   4.0   & 65.6   &  36.8   & 25.2   &  22.4  \\
62B & PaLM &  69.2   &  70.8   & 59.4   &  54.5   & 39.2   &  58.8   & 52.8   &  54.0   & 19.2   &   3.2   & 53.2   &  54.0   & 34.4   &   9.6   & 48.4   &  72.8   & 24.8   &  26.0    \\\vspace{3mm} 
& Flan-PaLM &  66.8   &  73.6   & 64.2   &  62.6   & 42.8   &  54.4   & 69.2   &  39.2   & 13.2   &   0.0   & 55.6   &  49.2   & 18.0   &  13.2   & 74.4   &  59.2   & 54.0   &  42.8   \\
540B & \palm{} &  83.2   &  80.0   & 61.0   &  59.4   & 53.6   &  79.2   & 60.8   &  67.6   & 28.4   &  28.0   & 53.6   &  51.2   & 37.6   &   0.0   & 70.8   &  90.4   & 39.6   &  49.2    \\
& Flan-PaLM&  86.0   &  83.2   & 65.2   &  63.1   & 58.0   &  74.0   & 76.8   &  69.6   & 29.2   &  23.6   & 62.4   &  52.8   & 40.0   &  43.6   & 67.6   &  88.8   & 54.4   &  52.4 \\
\midrule 
250M & Switch$_\textsc{base}$ & 0.0 &0.0 &2.7 &10.7 &0.0 &0.0 &0.0 &0.0 &0.0 &0.0 &0.0 &1.6 &0.0 &0.0 &0.0 &0.4 &0.0 &0.8\\\vspace{3mm} 
& \sshortname{}-Switch$_\textsc{base}$ & 51.2 &42.8 &55.1 &55.6 &18.8 &18.4 &63.6 &53.6 &0.0 &0.0 &56.8 &54.8 &9.6 &8.8 &64.8 &62.0 &34.8 &22.0 \\
780M & Switch$_\textsc{large}$  & 0.0 &26.0 &5.3 &5.3 &0.0 &10.8 &0.0 &0.0 &0.0 &0.0 &0.0 &15.2 &0.0 &8.4 &0.0 &48.4 &0.0 &0.0\\\vspace{3mm} 
& \sshortname{}-Switch$_\textsc{large}$  & 54.0 &22.0 &56.7 &50.8 &25.2 &24.0 &67.2 &59.2 &0.8 &0.0 &54.8 &43.6 &11.6 &3.6 &56.8 &30.0 &47.2 &28.0 \\
11B & Switch$_\textsc{xxl}$ & 0.0 &3.2 &0.0 &37.4 &0.0 &2.4 &0.0 &8.8 &0.0 &0.0 &0.0 &21.6 &0.0 &0.4 &0.0 &30.4 &0.0 &0.4\\
& \sshortname{}-Switch$_\textsc{xxl}$  & 56.2 &57.3 &65.5 &61.4 &60.9 &55.3 &70.4 &66.4 &0.8 &0.4 &57.3 &47.7 &12.8 &8.8 &58.1 &58.0 &61.2 &54.9 \\
\midrule
80M  & \sshortname{}-GS$_\textsc{small}$ & 60.0 &46.0 &51.9 &50.8 &21.2 &21.6 &30.4 &28.4 &1.2 &0.0 &54.8 &35.2 &9.6 &12.4 &56.0 &0.0 &21.6 &16.4 \\
250M & \sshortname{}-GS$_\textsc{base}$ & 48.0 &34.0 &53.5 &51.9 &27.6 &11.2 &65.2 &26.0 &0.0 &0.0 &53.2 &51.6 &9.6 &18.4 &59.6 &1.2 &35.6 &20.4
 \\
780M & \sshortname{}-GS$_\textsc{large}$ & 46.8 &41.2 &53.5 &50.8 &5.6 &37.2 &68.8 &66.0 &2.0 &0.0 &51.2 &12.4 &19.2 &12.8 &54.0 &50.8 &47.6 &28.4 \\
\midrule
80M  & \sshortname{}-EC$_\textsc{small}$  & 59.6 &39.2 &49.7 &53.5 &21.6 &17.2 &34.0 &36.4 &1.2 &0.0 &54.4 &45.6 &9.6 &0.4 &58.0 &0.4 &20.4 &23.2 \\
250M & \sshortname{}-EC$_\textsc{base}$ & 57.6 &43.6 &50.3 &50.8 &34.4 &24.8 &67.6 &34.4 &0.8 &0.0 &53.6 &17.2 &9.6 &7.6 &72.0 &44.0 &33.6 &24.0 \\
780M & \sshortname{}-EC$_\textsc{large}$ & 58.8 &48.0 &58.8 &50.8 &35.6 &43.2 &69.2 &70.0 &0.0 &0.0 &53.2 &30.8 &4.8 &5.6 &68.4 &52.8 &41.6 &21.6 \\
3B & \sshortname{}-EC$_\textsc{xl}$ & 54.3 &49.7 &59.9 &56.2 &48.4 &37.4 &69.0 &32.9 &-1.3 &0.4 &53.0 &50.0 &9.9 &4.0 &61.2 &40.1 &50.4 &38.9\\
\midrule
250M & ST$_\textsc{base}$ & 0.0 &9.2 &0.0 &35.8 &0.0 &14.4 &0.0 &0.8 &0.0 &0.0 &0.0 &52.8 &0.0 &0.0 &0.0 &0.4 &0.0 &18.8\\\vspace{3mm} 
& \sshortname{}-ST$_\textsc{base}$ & 48.0 &49.3 &59.6 &54.1 &11.6 &36.1 &66.1 &64.2 &1.0 &0.0 &50.0 &44.2 &19.5 &12.1 &51.4 &49.9 &49.6 &21.4 \\
32B & ST$_\textsc{32B}$ & 0.0 &0.0 &0.0 &0.0 &0.0 &32.8 &0.0 &0.4 &0.0 &0.0 &0.0 &0.0 &0.0 &1.2 &0.0 &0.4 &0.0 &6.4\\
& \sshortname{}-ST$_\textsc{32B}$ & 63.6 &67.6 &67.9 &65.8 &66.4 &62.0 &70.8 &74.8 &15.2 &0.0 &58.8 &42.0 &22.8 &5.2 &60.0 &54.4 &64.0 &49.6\\
  \bottomrule
\end{tabular}%
}
\end{table}

\begin{table}[]
\centering
\caption{BBH[9:18] individual task performance.}
\label{tab:my-table}
\setlength{\tabcolsep}{3pt}
\resizebox{\columnwidth}{!}{%
\begin{tabular}{llcccccccccccccccccc}
\toprule
& & \multicolumn{18}{c}{BBH} \\
\cmidrule(lr){3-20}
 &
  & \multicolumn{2}{c}{\thead{Logical Deduction\\Seven Objects}} &
  \multicolumn{2}{c}{\thead{Logical Deduction\\Three Objects}} &
  \multicolumn{2}{c}{\thead{Movie\\Recommendation}} &
  \multicolumn{2}{c}{\thead{Multistep\\Arithmetic}} &
  \multicolumn{2}{c}{\thead{Navigate}} &
  \multicolumn{2}{c}{\thead{Object\\Counting}} &
  \multicolumn{2}{c}{\thead{Penguins\\in a Table}} &
  \multicolumn{2}{c}{\thead{Reasoning about\\Colored Objects}} &
  \multicolumn{2}{c}{\thead{Ruin\\Names}} \\
  \cmidrule(lr){3-4}
  \cmidrule(lr){5-6}
  \cmidrule(lr){7-8}
  \cmidrule(lr){9-10}
  \cmidrule(lr){11-12}
  \cmidrule(lr){13-14}
  \cmidrule(lr){15-16}
  \cmidrule(lr){17-18}
  \cmidrule(lr){19-20}
 \multicolumn{2}{l}{Model} 
  & \thead{Direct} &
  \thead{CoT} &
  \thead{Direct} &
  \thead{CoT} &
  \thead{Direct} &
  \thead{CoT} &
  \thead{Direct} &
  \thead{CoT} &
  \thead{Direct} &
  \thead{CoT} &
  \thead{Direct} &
  \thead{CoT} &
  \thead{Direct} &
  \thead{CoT} &
  \thead{Direct} &
  \thead{CoT} &
  \thead{Direct} &
  \thead{CoT} \\
  \midrule
- & davinci & 20.0 & 27.2 & 38.0 & 52.0 & 58.8 & 71.2 & 0.8 & 1.6 & 58.0 & 66.0 & 33.2 & 49.6 & 28.1 & 35.6 & 13.2 & 41.2 & 18.4 & 33.2 \\
- & text-davinci-002 & 26.8 & 38.0 & 45.2 & 87.6 & 72.0 & 78.8 & 1.2 & 53.2 & 68.0 & 88.8 &  44.0 & 77.2 & 47.3 & 81.5 & 47.6 & 78.4 & 65.6 & 62.8 \\
- & text-davinci-003 & 40.0 & 52.4 & 62.0 & 88.0 & 79.2 & 83.6 & 1.2 & 49.6 & 53.2 & 94.4 & 33.2 & 82.0 & 52.1 & 83.6 & 67.2 & 86.8 & 82.0 & 58.8 \\\vspace{3mm}
- & code-davinci-002 & 26.0 & 38.8 & 52.8 & 87.6 & 84.8 & 90.4 & 1.2 & 47.6 & 50.4 & 96.4 & 45.2 & 93.2 & 66.4 & 79.5 & 67.6 & 91.6 & 75.2 & 68.4 \\
80M & T5-Small & 13.2   &   5.2   & 31.6   &  14.0   & 26.0   &  14.8   &  0.0   &   0.0   & 55.2   &  40.0   & 10.0   &   0.0   & 21.9   &  19.2   & 16.0   &  11.2   & 22.4   &   1.6    \\\vspace{3mm} 
& Flan-T5-Small &  16.8   &  11.2   & 30.8   &  30.0   & 43.2   &  20.4   &  0.0   &   1.6   & 58.0   &  58.0   &  5.6   &   3.2   & 21.9   &  10.3   & 17.2   &  10.8   & 13.2   &   0.8    \\
250M & T5-Base & 14.8   &   2.4   & 29.6   &  22.4   & 27.6   &   0.4   &  0.4   &   0.0   & 48.0   &  42.0   &  8.8   &   0.0   & 21.9   &  19.2   & 15.6   &  12.4   & 28.0   &   2.4    \\\vspace{3mm} 
& Flan-T5-Base &  24.4 &19.2 &42.8 &40.8 &39.6 &32.4 &0.4 &0.0 &62.8 &32.4 &22.8 &11.2 &17.8 &9.6 &22.4 &23.6 &13.6 &10.4   \\
780M & T5-Large &  13.2   &   8.0   & 32.4   &  26.0   & 24.8   &  23.2   &  0.4   &   0.0   & 42.0   &  42.0   &  9.6   &   6.4   & 21.9   &  23.3   & 10.4   &  14.8   & 27.6   &   0.4   \\\vspace{3mm} 
& Flan-T5-Large &  46.8 &22.4 &53.2 &36.8 &41.6 &28.0 &0.4 &0.4 &44.8 &34.0 &32.8 &16.8 &22.6 &22.6 &43.6 &38.4 &28.8 &25.6  \\
3B & T5-XL &  13.6   &  15.2   & 35.2   &  35.6   & 25.2   &  23.6   &  0.8   &   0.8   & 42.0   &  38.0   &  6.4   &  25.2   & 21.2   &  25.3   & 12.8   &  14.8   & 26.0   &   0.8   \\\vspace{3mm} 
& Flan-T5-XL &  53.6 &25.2 &66.0 &50.8 &46.4 &36.4 &0.4 &0.4 &48.4 &46.4 &42.4 &30.8 &37.7 &35.6 &50.8 &46.0 &42.0 &28.4
   \\
11B & T5-XXL  &  18.0   &  18.0   & 36.8   &  42.8   & 46.0   &  45.2   &  0.0   &   0.0   & 41.6   &  37.2   & 31.6   &  33.2   & 21.2   &  24.7   & 16.4   &  22.8   & 20.8   &   0.0   \\\vspace{3mm} 
& Flan-T5-XXL & 54.8 &48.8 &76.0 &58.8 &53.2 &53.2 &0.4 &0.4 &60.4 &54.0 &50.8 &34.0 &39.0 &39.0 &58.8 &46.8 &52.4 &53.2   \\
8B & PaLM &  13.2   &  14.8   & 35.6   &  36.4   & 28.4   &  26.4   &  0.8   &   1.2   & 58.0   &  58.0   & 36.8   &  18.8   & 25.3   &  19.9   & 18.0   &  18.8   & 21.2   &  24.4   \\\vspace{3mm} 
& Flan-PaLM &  25.6   &  12.8   & 47.6   &  40.8   & 72.8   &  43.6   &  0.8   &   0.8   & 58.4   &  55.6   & 30.0   &  24.8   & 26.7   &  30.1   & 28.4   &  34.0   & 36.8   &  32.0   \\
62B & PaLM &  19.6   &  20.0   & 36.8   &  52.4   & 60.8   &  70.8   &  0.8   &   1.6   & 56.4   &  55.2   & 41.6   &  50.4   & 24.0   &  37.0   & 17.2   &  48.0   & 50.4   &  54.0   \\\vspace{3mm} 
& Flan-PaLM &  48.8   &  34.0   & 74.0   &  56.0   & 82.0   &  72.8   &  1.2   &   1.6   & 60.4   &  49.2   & 50.4   &  51.2   & 37.0   &  49.3   & 50.4   &  46.0   & 63.6   &  54.8    \\
540B & \palm{} &  24.8   &  43.6   & 63.6   &  78.0   & 87.2   &  92.0   &  1.6   &  19.6   & 62.4   &  79.6   & 51.2   &  83.2   & 44.5   &  65.1   & 38.0   &  74.4   & 76.0   &  61.6    \\
& Flan-PaLM &  50.8   &  48.4   & 85.6   &  87.2   & 85.6   &  82.4   &  0.8   &  29.6   & 68.4   &  78.0   & 54.0   &  88.8   & 55.5   &  72.6   & 66.4   &  82.4   & 81.2   &  68.0 \\
\midrule 
250M & Switch$_\textsc{base}$ & 0.0 &0.4 &0.0 &1.2 &0.0 &3.6 &0.4 &0.0 &0.0 &0.0 &0.0 &0.0 &0.0 &0.0 &0.0 &6.4 &0.0 &0.0 \\\vspace{3mm}
& \sshortname{}-Switch$_\textsc{base}$ & 38.4 &23.2 &47.2 &41.6 &41.6 &33.2 &0.0 &0.0 &59.2 &54.0 &30.8 &18.4 &34.9 &19.9 &36.8 &24.8 &12.4 &10.4 \\
780M & Switch$_\textsc{large}$ & 0.0 &0.0 &0.0 &0.0 &0.0 &0.0 &0.0 &0.4 &0.0 &0.0 &0.4 &0.0 &0.0 &17.8 &0.0 &4.0 &0.0 &0.4\\\vspace{3mm}
& \sshortname{}-Switch$_\textsc{large}$ & 44.8 &22.8 &57.2 &42.0 &61.2 &47.2 &0.4 &0.8 &45.6 &43.2 &41.6 &33.2 &38.4 &29.5 &42.0 &32.4 &11.6 &10.8 \\
11B & Switch$_\textsc{xxl}$ & 0.0 &0.0 &0.0 &4.0 &0.0 &1.2 &0.4 &0.0 &0.0 &0.0 &0.0 &1.6 &0.0 &6.8 &0.0 &2.0 &0.0 &2.0\\
& \sshortname{}-Switch$_\textsc{xxl}$ & 61.1 &46.9 &80.6 &70.6 &58.5 &54.1 &1.5 &0.4 &58.4 &58.2 &47.2 &40.3 &47.6 &44.2 &62.8 &55.7 &66.4 &50.4 \\
\midrule
80M  & \sshortname{}-GS$_\textsc{small}$ & 16.8 &12.4 &33.6 &34.4 &42.8 &13.2 &0.0 &0.4 &62.4 &40.0 &20.0 &9.2 &13.0 &15.8 &25.6 &19.2 &9.2 &6.4 \\
250M & \sshortname{}-GS$_\textsc{base}$ & 36.0 &17.2 &48.4 &35.6 &54.0 &47.2 &0.0 &0.0 &61.2 &53.6 &27.2 &29.6 &29.5 &20.5 &34.0 &24.4 &10.8 &14.0 \\
780M & \sshortname{}-GS$_\textsc{large}$ & 46.8 &26.0 &60.8 &34.4 &45.2 &39.6 &1.6 &0.4 &57.6 &44.8 &36.0 &21.6 &31.5 &25.3 &25.6 &32.4 &29.6 &32.4 \\
\midrule
80M  & \sshortname{}-EC$_\textsc{small}$ & 14.8 &12.8 &33.6 &29.6 &40.4 &36.0 &0.8 &0.4 &64.4 &57.6 &19.6 &4.0 &13.7 &17.8 &21.6 &18.8 &8.8 &8.0 \\
250M & \sshortname{}-EC$_\textsc{base}$ & 35.2 &24.0 &50.8 &34.8 &24.8 &34.0 &0.4 &0.4 &62.0 &50.4 &32.8 &24.8 &31.5 &26.0 &33.2 &26.0 &18.0 &15.2 \\
780M & \sshortname{}-EC$_\textsc{large}$ & 50.0 &22.8 &57.2 &30.0 &50.8 &45.2 &0.0 &0.8 &58.8 &59.6 &38.4 &31.2 &33.6 &27.4 &34.4 &39.6 &20.0 &26.4
 \\
3B & \sshortname{}-EC$_\textsc{xl}$  & 53.4 &48.6 &60.8 &56.5 &48.6 &38.4 &66.7 &35.1 &0.0 &0.4 &53.6 &49.2 &11.0 &4.5 &61.4 &40.3 &53.0 &37.9 \\
\midrule
250M & ST$_\textsc{base}$ & 0.0 &13.2 &0.0 &28.8 &0.0 &4.0 &0.0 &1.6 &0.0 &42.0 &0.0 &6.4 &0.0 &15.8 &0.0 &6.4 &0.0 &0.8 \\\vspace{3mm}
& \sshortname{}-ST$_\textsc{base}$ & 43.5 &22.7 &53.7 &42.6 &42.9 &33.9 &0.4 &0.4 &48.1 &47.2 &33.1 &31.6 &35.0 &27.7 &40.0 &40.7 &18.9 &21.0 \\
32B & ST$_\textsc{32B}$ & 0.0 &1.6 &0.0 &20.8 &0.0 &0.4 &0.4 &0.4 &0.0 &0.0 &0.4 &3.2 &0.0 &0.0 &0.0 &10.4 &0.0 &0.0\\
& \sshortname{}-ST$_\textsc{32B}$ & 62.4 &44.8 &90.8 &79.6 &69.6 &66.0 &0.8 &0.4 &63.2 &48.0 &52.4 &49.6 &61.6 &55.5 &78.0 &72.0 &72.8 &64.4 \\
  \bottomrule
\end{tabular}%
}
\end{table}

\clearpage
\begin{table}[]
\centering
\caption{BBH[18:27] individual task performance.}
\label{tab:my-table}
\setlength{\tabcolsep}{3pt}
\resizebox{\columnwidth}{!}{%
\begin{tabular}{llcccccccccccccccccccc}
\toprule
& & \multicolumn{19}{c}{BBH}  \\
\cmidrule(lr){3-22}
 &
  & \multicolumn{2}{c}{\thead{Salient Translation\\Error Detection}} &
  \multicolumn{2}{c}{\thead{Snarks}} &
  \multicolumn{2}{c}{\thead{Sports\\Understanding}} &
  \multicolumn{2}{c}{\thead{Temporal\\Sequences}} &
  \multicolumn{2}{c}{\thead{Tracking Shuffled\\Objects (5)}} &
  \multicolumn{2}{c}{\thead{Tracking Shuffled\\Objects (7)}} &
  \multicolumn{2}{c}{\thead{Tracking Shuffled\\Objects (3)}} &
  \multicolumn{2}{c}{\thead{Web of\\Lies}} &
  \multicolumn{2}{c}{\thead{Word\\Sorting}} &
  \multicolumn{2}{c}{\thead{\textbf{Average}}} \\
  \cmidrule(lr){3-4}
  \cmidrule(lr){5-6}
  \cmidrule(lr){7-8}
  \cmidrule(lr){9-10}
  \cmidrule(lr){11-12}
  \cmidrule(lr){13-14}
  \cmidrule(lr){15-16}
  \cmidrule(lr){17-18}
  \cmidrule(lr){19-20}
  \cmidrule(lr){21-22}
 \multicolumn{2}{l}{Model} 
  & \thead{Direct} &
  \thead{CoT} &
  \thead{Direct} &
  \thead{CoT} &
  \thead{Direct} &
  \thead{CoT} &
  \thead{Direct} &
  \thead{CoT} &
  \thead{Direct} &
  \thead{CoT} &
  \thead{Direct} &
  \thead{CoT} &
  \thead{Direct} &
  \thead{CoT} &
  \thead{Direct} &
  \thead{CoT} &
  \thead{Direct} &
  \thead{CoT} &
  \thead{Direct} &
  \thead{CoT} \\
  \midrule
- & davinci & 22.4 & 5.2 & 52.2 & 47.8 & 54.4 & 94.0 & 22.8 & 22.4 & 32.0 & 18.0 & 13.6 & 14.8 & 33.6 & 32.0 & 48.8 & 59.2 & 11.2 & 6.0 & 33.6 & 38.3 \\
- & text-davinci-002 & 61.6 & 62.4 & 65.2 & 60.7 & 71.6 & 92.0 & 33.6 & 67.2 & 23.2 & 60.8 & 17.2 & 59.6 & 34.8 & 62.8 & 51.6 & 92.0 & 36.8 & 44.4 & 48.6 & 67.2 \\
- & text-davinci-003 & 68.0 & 60.8 & 67.4 & 74.2 & 72.4 & 96.0 & 37.6 & 58.0 & 18.0 & 80.8 & 16.0 & 81.2 & 30.4 & 68.4 & 53.2 & 100.0 & 45.6 & 41.6 & 50.9 & 70.7 \\\vspace{3mm}
- & code-davinci-002 & 62.0 & 60.8 & 61.2 & 59.6 & 72.8 & 97.6 & 77.6 & 96.8 & 20.4 & 89.6 & 14.4 & 85.6 & 37.6 & 78.4 & 51.6 & 95.2 & 50.4 & 40.4 & 52.8 & 73.7\\
80M & T5-Small &  12.0   &   0.0   & 46.1   &  15.2   & 46.4   &  35.6   & 28.4   &   1.6   & 20.8   &   0.0   & 15.2   &   0.0   & 32.8   &   0.0   & 51.2   &   0.0   &  0.4   &   0.0  & 27.0 & 7.2 \\ \vspace{3mm} 
& Flan-T5-Small &  22.4   &  15.2   & 46.6   &   9.6   & 54.8   &  54.0   & 28.4   &  17.2   & 22.4   &  15.2   & 14.0   &   8.8   & 30.8   &  25.6   & 53.6   &  36.8   &  2.0   &   1.2  & 29.1 & 19.2 \\
250M & T5-Base &  22.0   &   0.8   & 46.1   &   5.1   & 46.4   &  38.4   & 28.4   &  28.4   & 20.4   &   5.6   & 15.2   &   5.6   & 31.6   &   9.6   & 51.6   &  22.4   &  0.8   &   3.2   & 27.8 & 14.6 \\\vspace{3mm} 
& Flan-T5-Base &  11.6 &18.0 &42.7 &46.1 &52.8 &46.4 &18.4 &20.4 &16.8 &19.2 &10.4 &11.2 &33.2 &32.0 &52.4 &47.2 &4.0 &2.0 &30.3 &26.8 \\
780M & T5-Large &  22.4   &   0.0   & 46.1   &  14.6   & 46.8   &  48.4   & 28.0   &  28.4   & 22.0   &  16.4   & 15.2   &   9.2   & 32.0   &  22.8   & 49.2   &  22.8   &  3.2   &   0.0 & 27.7 & 16.1  \\\vspace{3mm} 
& Flan-T5-Large & 41.6 &25.6 &57.9 &52.8 &52.0 &45.2 &8.4 &23.2 &12.4 &11.2 &8.4 &10.4 &33.6 &31.6 &51.2 &48.4 &0.8 &2.4 &34.7 &28.5 \\
3B & T5-XL &  22.8   &   6.8   & 47.2   &  30.3   & 50.8   &  44.8   & 28.4   &  22.8   & 15.2   &  14.8   & 12.4   &  12.0   & 32.4   &  31.2   & 48.8   &  43.2   &  2.4   &   2.4   & 27.4 & 19.2   \\\vspace{3mm} 
& Flan-T5-XL &  34.4 &30.4 &72.5 &75.8 &51.2 &55.6 &22.8 &31.2 &12.4 &15.6 &8.4 &10.0 &29.2 &29.6 &49.6 &46.8 &4.8 &0.0 &40.2 &35.9 \\
11B & T5-XXL  &  15.2   &   0.0   & 53.9   &  25.3   & 47.2   &  60.0   & 19.2   &  17.2   & 18.4   &   1.6   & 10.0   &   0.0   & 33.2   &  30.0   & 48.8   &   4.4   &  3.2   &   2.0  & 29.5 & 19.3 \\\vspace{3mm} 
& Flan-T5-XXL &  46.4 &50.0 &74.7 &76.4 &64.4 &66.0 &25.6 &21.2 &18.0 &12.0 &9.6 &16.8 &28.8 &24.8 &54.0 &53.2 &7.2 &4.4 &45.6 &41.6 \\
8B & PaLM &  21.6   &  12.0   & 53.9   &  51.1   & 54.0   &  76.8   & 25.6   &  28.8   & 20.4   &  19.6   & 12.8   &  10.8   & 32.0   &  31.6   & 51.2   &  48.8   &  4.4   &   4.4 & 30.8 & 30.1  \\\vspace{3mm} 
& Flan-PaLM &  23.2   &   0.8   & 69.1   &  59.6   & 64.4   &  69.6   & 15.6   &  24.0   & 17.2   &  11.2   & 16.8   &  13.6   & 33.2   &  32.0   & 52.0   &  49.2   &  6.0   &   1.2 & 36.4 & 31.1  \\
62B & PaLM &  28.0   &  21.6   & 52.8   &  48.3   & 78.4   &  95.6   & 21.2   &  26.4   & 19.6   &  18.8   & 13.6   &  13.6   & 30.4   &  36.4   & 48.8   &  80.8   &  7.6   &   8.4 & 37.4 & 42.3   \\\vspace{3mm} 
& Flan-PaLM &  45.2   &  40.4   & 83.1   &  78.1   & 79.2   &  81.2   & 30.8   &  36.0   & 21.2   &  18.0   & 15.2   &  18.0   & 22.0   &  29.6   & 48.4   &  92.0   & 11.2   &  10.0  & 47.5 & 44.9  \\
540B & \palm{} &  48.8   &  54.0   & 78.1   &  61.8   & 80.4   &  98.0   & 39.6   &  78.8   & 16.8   &  57.6   & 13.6   &  42.4   & 28.4   &  58.8   & 51.2   &  100.0  & 32.0   &  21.6  & 49.1 & 62.0 \\
& Flan-PaLM &  53.2   &  51.6   & 85.4   &  76.4   & 83.2   &  87.2   & 81.6   &  91.6   & 24.4   &  50.8   & 21.6   &  38.0   & 32.4   &  71.6   & 62.4   &  100.0  & 32.0   &  33.2 & 57.9 & 66.3  \\
\midrule 
250M & Switch$_\textsc{base}$ & 0.0 &0.0 &0.0 &0.0 &0.0 &0.0 &0.0 &13.6 &0.0 &0.0 &0.0 &0.0 &0.0 &0.0 &0.0 &0.0 &0.0 &0.0 &0.1 &1.4\\\vspace{3mm}
& \sshortname{}-Switch$_\textsc{base}$ & 27.2 &25.6 &39.3 &39.9 &53.2 &54.4 &10.4 &15.6 &11.6 &13.2 &14.4 &14.4 &32.0 &33.6 &49.6 &53.2 &2.4 &1.2 &33.2 &29.4 \\
780M & Switch$_\textsc{large}$ & 0.0 &0.4 &0.0 &45.5 &0.0 &0.0 &0.0 &6.4 &0.0 &0.0 &0.0 &0.0 &0.0 &0.0 &0.0 &4.0 &0.0 &0.0 &0.2 &7.2 \\\vspace{3mm}
& \sshortname{}-Switch$_\textsc{large}$ & 27.6 &8.8 &52.8 &52.8 &57.2 &54.4 &18.4 &14.8 &12.4 &12.8 &8.4 &10.8 &33.6 &30.4 &51.2 &48.0 &4.0 &0.4 &36.4 &28.0 \\
11B & Switch$_\textsc{xxl}$ & 0.0 &6.8 &0.0 &0.0 &0.0 &12.0 &0.0 &0.0 &0.0 &0.0 &0.0 &0.0 &0.0 &0.0 &0.0 &39.6 &0.0 &0.0 &0.0 &6.7 \\
& \sshortname{}-Switch$_\textsc{xxl}$ & 51.7 &41.1 &81.1 &74.3 &68.8 &74.3 &40.0 &36.4 &19.5 &18.0 &21.0 &14.0 &20.8 &25.7 &50.3 &49.7 &8.3 &4.7 &47.9 &43.4 \\
\midrule
80M  & \sshortname{}-GS$_\textsc{small}$ & 20.8 &0.0 &46.6 &37.1 &54.0 &52.8 &22.4 &22.4 &23.6 &18.0 &12.4 &8.8 &34.4 &32.0 &51.6 &32.0 &2.4 &0.0 &29.6 &20.9 \\
250M & \sshortname{}-GS$_\textsc{base}$ & 23.2 &0.0 &47.8 &35.4 &56.4 &52.8 &22.8 &19.2 &12.4 &15.6 &8.4 &10.8 &32.4 &34.8 &50.0 &52.8 &3.6 &0.4 &33.7 &25.1 \\
780M & \sshortname{}-GS$_\textsc{large}$ & 16.8 &14.8 &61.8 &53.9 &59.2 &55.2 &12.4 &20.8 &12.4 &5.6 &8.4 &5.6 &34.0 &19.2 &52.4 &56.0 &3.2 &1.6 &35.0 &29.2 \\
\midrule
80M  & \sshortname{}-EC$_\textsc{small}$ & 23.2 &3.6 &48.3 &23.6 &54.0 &54.4 &17.6 &23.6 &24.8 &18.8 &11.6 &14.0 &30.0 &28.8 &50.8 &30.8 &2.8 &0.0 &29.2 &22.2 \\
250M & \sshortname{}-EC$_\textsc{base}$ & 22.4 &13.2 &41.6 &44.4 &57.2 &54.0 &16.0 &11.2 &14.4 &14.8 &8.0 &10.0 &34.0 &34.0 &53.2 &52.4 &2.8 &1.2 &34.0 &26.6 \\
780M & \sshortname{}-EC$_\textsc{large}$ & 42.0 &15.6 &55.6 &56.7 &59.2 &58.4 &19.6 &20.8 &12.4 &12.8 &8.4 &9.2 &33.6 &32.0 &54.4 &49.2 &3.6 &2.8 &37.9 &32.0 \\
3B & \sshortname{}-EC$_\textsc{xl}$ &38.6 &21.2 &64.0 &53.7 &63.2 &59.2 &16.6 &22.4 &13.2 &17.0 &8.6 &8.6 &26.8 &28.1 &50.8 &48.8 &6.8 &2.3 &40.3 &33.2 \\
\midrule
250M & ST$_\textsc{base}$ & 0.0 &10.8 &0.0 &44.4 &0.0 &47.2 &0.0 &2.0 &0.0 &0.0 &0.0 &0.0 &0.0 &0.0 &0.0 &21.2 &0.0 &0.0 &0.0 &14.0 \\\vspace{3mm}
& \sshortname{}-ST$_\textsc{base}$ & 13.3 &11.6 &61.0 &58.1 &56.0 &52.2 &18.4 &20.2 &12.2 &12.3 &7.9 &12.2 &33.9 &34.5 &52.5 &48.6 &3.3 &2.2 &34.7 &26.6 \\
32B & ST$_\textsc{32B}$ & 0.0 &10.4 &0.0 &0.0 &0.0 &0.0 &0.0 &0.4 &0.0 &18.0 &0.0 &9.2 &0.0 &32.8 &0.0 &0.0 &0.0 &0.0 &0.0 &5.5 \\
& \sshortname{}-ST$_\textsc{32B}$ & 57.6 &52.8 &88.2 &86.0 &73.2 &75.6 &75.6 &44.8 &27.2 &18.4 &28.0 &19.6 &21.6 &28.0 &40.4 &48.8 &15.6 &4.8 &54.4 &47.4 \\
  \bottomrule
\end{tabular}%
}
\end{table}

\subsection{Reasoning}

\begin{table}[]
\centering
\caption{Reasoning[:4] individual task performance.}
\label{tab:my-table}
{%
\begin{tabular}{llcccccccccc}
\toprule
& & \multicolumn{5}{c}{Reasoning}  \\
\cmidrule(lr){3-7}
 &
  & \multicolumn{1}{c}{\thead{GSM8K}} &
  \multicolumn{1}{c}{\thead{ASDIV}} &
  \multicolumn{1}{c}{\thead{StrategyQA}} &
  \multicolumn{1}{c}{\thead{SVAMP}} &
  \multicolumn{1}{c}{\thead{\textbf{Average}}} \\
  \cmidrule(lr){3-3}
  \cmidrule(lr){4-4}
  \cmidrule(lr){5-5}
  \cmidrule(lr){6-6}
  \cmidrule(lr){7-7}
 \multicolumn{2}{l}{Model} &  
  \thead{CoT} &
  \thead{CoT} &
  \thead{CoT} &
  \thead{CoT} &
  \thead{CoT} &
\\
  \midrule
80M & T5-Small & 1.1 & 1.7 & 37.1 & 1.3 & 10.3 \\ \vspace{3mm}
& Flan-T5-Small & 2.1 & 2.8 & 53.2 & 2.1 & 15.0\\ 
250M & T5-Base & 2.0 & 1.8 & 52.8 & 2.0 & 14.7 \\ \vspace{3mm}
& Flan-T5-Base & 3.9 & 4.9 & 53.3 & 3.5 & 16.4\\
780M & T5-Large & 1.6 & 2.0 & 42.8 & 1.0 & 11.9 \\ \vspace{3mm}
& Flan-T5-Large & 8.6 & 14.5 & 54.2 & 11.6 & 22.2 \\
3B & T5-XL & 2.7 & 5.2 & 45.9 & 2.9 & 14.2 \\ \vspace{3mm}
& Flan-T5-XL & 16.9 & 28.2 & 64.6 & 25.9 & 33.9 \\
11B & T5-XXL & 2.5 & 15.0 & 55.0 & 12.9 & 21.4 \\ \vspace{3mm}
& Flan-T5-XXL & 26.7 & 47.4 & 69.9 & 41.4 & 46.3 \\
8B & Flan-PaLM & 21.4 & 37.5 & 65.5 & 23.1 & 36.9 \\
62B & Flan-PaLM  & 47.5 & 64.5 & 76.4 & 50.2 & 47.7 \\
540B & Flan-PaLM & 73.0 & 77.7 & 83.0 & 72.2 & 76.5 \\
\midrule 
250M & Switch$_\textsc{base}$ & 0.6 & 1.0 & 17.5 & 1.5 & 5.2 \\\vspace{3mm}
& \sshortname{}-Switch$_\textsc{base}$ & 6.4 & 8.4 & 53.3 & 6.3 & 18.6 \\
780M & Switch$_\textsc{large}$ & 1.9 & 2.4 & 43.2 & 2.0 & 12.4 \\\vspace{3mm}
& \sshortname{}-Switch$_\textsc{large}$ & 12.7 & 19.0 & 56.3 & 13.0 & 25.3  \\
11B & Switch$_\textsc{xxl}$ & 0.2 & 0.4 & 36.2 & 0.1 & 9.2 \\
& \sshortname{}-Switch$_\textsc{xxl}$ & 27.0 & 47.8 & 70.1 & 41.7 & 46.6 \\
\midrule
80M  & \sshortname{}-GS$_\textsc{small}$ & 3.7 & 5.0 & 53.3 & 3.3 & 16.1 \\
250M & \sshortname{}-GS$_\textsc{base}$ & 11.1 & 13.9 &  53.7 & 9.9 & 22.2 \\
780M & \sshortname{}-GS$_\textsc{large}$ &16.7 & 22.2 & 54.6 & 17.0 & 27.6  \\
\midrule
80M  & \sshortname{}-EC$_\textsc{small}$ & 5.2 & 5.6 & 53.3 & 5.4 & 16.6 \\
250M & \sshortname{}-EC$_\textsc{base}$ & 10.7 & 13.7 & 53.3 & 10.5 & 22.0 \\
780M & \sshortname{}-EC$_\textsc{large}$ & 15.9 & 25.7 & 65.5 & 21.7 & 32.2 \\
3B & \sshortname{}-EC$_\textsc{xl}$ & 21.3 & 33.6 & 67.2 & 30.3 &  38.1 \\
\midrule
250M & ST$_\textsc{base}$ & 2.0 & 1.9 & 45.0 & 1.3 & 12.6 \\\vspace{3mm}
& \sshortname{}-ST$_\textsc{base}$ & 11.2 & 11.1 & 59.8 & 8.0 &  22.5 \\
& ST$_\textsc{32B}$ & 2.7 & 18.4 & 1.7 & 16.2 & 9.8 \\
& \sshortname{}-ST$_\textsc{32B}$ & 51.1 & 65.3 & 80.8 & 68.1 & 66.3\\
  \bottomrule
\end{tabular}%
}
\end{table}
The four reasoning tasks are held-in, which means we perform instruction finetuning on the training set while evaluating on the ``validation'' set in a few-shot way. 
The detailed performance is presented here. 

\clearpage
\subsection{QA}
We perform evaluation on four held-out QA tasks and the results are summarized in this section.  

\begin{table}[]
\centering
\caption{QA[:5] individual task performance.}
\label{tab:my-table}
{%
\begin{tabular}{llcccccccccc}
\toprule
& & \multicolumn{5}{c}{QA}  \\
\cmidrule(lr){3-7}
 &
  & \multicolumn{1}{c}{\thead{UnifiedQA\\Elementary Science}} &
  \multicolumn{1}{c}{\thead{ARC\\easy}} &
  \multicolumn{1}{c}{\thead{ARC\\challlenge}} &
  \multicolumn{1}{c}{\thead{BoolQ}} &
  \multicolumn{1}{c}{\thead{\textbf{Average}}} \\
  \cmidrule(lr){3-3}
  \cmidrule(lr){4-4}
  \cmidrule(lr){5-5}
  \cmidrule(lr){6-6}
  \cmidrule(lr){7-7}
 \multicolumn{2}{l}{Model} 
  & \thead{Direct} &
  \thead{Direct} &
  \thead{Direct} &
  \thead{Direct} &
  \thead{Direct} &
\\
  \midrule
80M & Flan-T5-Small & 27.6 & 40.4 & 31.9  & 63.7 & 40.9 \\ 
250M & Flan-T5-Base & 34.1 & 46.1 & 38.7 & 76.2 &  48.8\\
780M & Flan-T5-Large & 43.9 & 76.3 & 53.2 & 84.0 & 64.4 \\
3B & Flan-T5-XL & 53.7 & 88.4 & 66.2  & 88.0 & 74.1 \\
11B & Flan-T5-XXL & 63.4 & 94.2 & 74.6 & 89.3 &  80.4 \\
8B & Flan-PaLM & 72.4 &	83.4 & 61.7 & 83.0 & 75.1 \\
62B & Flan-PaLM & 85.4 & 92.0 & 77.3 & 86.3 & 85.3 \\
540B & Flan-PaLM & 92.7 & 95.2 & 88.7 & 83.0 & 89.9 \\
\midrule 
250M & \sshortname{}-Switch$_\textsc{base}$ & 48.1 & 61.4 &	43.2 & 79.3 & 58.0 \\
780M & \sshortname{}-Switch$_\textsc{large}$ & 50.3 & 70.3 & 61.7 & 83.8 & 66.5 \\
11B & \sshortname{}-Switch$_\textsc{xxl}$ & 60.2 & 73.7 & 91.7 & 89.7 &  78.8 \\
\midrule
80M  & \sshortname{}-GS$_\textsc{small}$   & 39.0 & 48.5 & 36.0 & 72.0 & 48.9 \\
250M & \sshortname{}-GS$_\textsc{base}$ & 43.9 & 59.3 & 45.9 & 82.5 &  57.9 \\
780M & \sshortname{}-GS$_\textsc{large}$ & 53.7 & 69.4 & 66.7 & 88.2 &  69.5 \\
\midrule
80M  & \sshortname{}-EC$_\textsc{small}$ & 37.4 & 61.4 & 50.0 & 83.4 &  58.1  \\
250M & \sshortname{}-EC$_\textsc{base}$ & 51.2 & 61.4 & 50.0 & 83.4 &   61.5 \\
780M & \sshortname{}-EC$_\textsc{large}$ & 59.3 & 71.8 & 71.3 & 90.1 &  73.1 \\
3B & \sshortname{}-EC$_\textsc{xl}$  & 60.1 & 71.8 & 75.3 & 90.1 & 74.3 \\
\midrule
250M & \sshortname{}-ST$_\textsc{base}$ & 47.2 & 58.3 & 57.7 & 82.6 & 61.5 \\
32B & ST$_\textsc{32B}$ & 31.7 & 25.8 & 30.1 & 40.6 & 32.1 \\
& \sshortname{}-ST$_\textsc{32B}$ & 69.9 & 99.2 & 90.8 & 92.1 & 88.0 \\
  \bottomrule
\end{tabular}%
}
\end{table}

\end{document}